\documentclass[11pt]{article}

\usepackage[preprint]{acl}

\usepackage{times}
\usepackage{latexsym}

\usepackage[T1]{fontenc}

\usepackage[utf8]{inputenc}

\usepackage{microtype}

\usepackage{inconsolata}

\usepackage{graphicx}

\usepackage{todonotes}
\usepackage{comment}
\usepackage{amsmath}
\usepackage{hyperref}
\usepackage{placeins}

\usepackage{url}
\usepackage{placeins}
\usepackage{xcolor}
\usepackage{listings}
\usepackage{caption}
\usepackage{amssymb}
\usepackage{upquote}

\usepackage{enumitem}

\usepackage{fontawesome5}
\usepackage{hyperref}
\usepackage{multirow}

\definecolor{bsnavy}{rgb}{0.02,0.10,0.27}
\definecolor{bsblue}{rgb}{0.08,0.30,0.71}
\definecolor{bsteal}{rgb}{0.03,0.51,0.53}
\definecolor{bscyan}{rgb}{0.12,0.63,0.69}
\definecolor{bspurple}{rgb}{0.30,0.16,0.75}
\definecolor{bslavender}{rgb}{0.55,0.46,0.90}
\definecolor{bsorange}{rgb}{0.96,0.59,0.49}
\definecolor{codegray}{rgb}{0.42,0.42,0.42}
\definecolor{codeink}{rgb}{0.10,0.12,0.15}
\definecolor{pathcolor}{rgb}{0.588,0,0.094}
\definecolor{backcolour}{rgb}{0.995,0.995,0.995}
\definecolor{pyframe}{rgb}{0.82,0.84,0.88}
\definecolor{bsframe}{rgb}{0.56,0.72,0.86}
\definecolor{bsgraphite}{HTML}{171922}

\definecolor{bggreen}{rgb}{0.945,0.980,0.953}
\definecolor{bordergreen}{rgb}{0.545,0.769,0.541}

\definecolor{listinggraphite}{HTML}{171922}
\definecolor{listingpurple}{HTML}{6657D9}
\definecolor{listingdarkpurple}{HTML}{453A9E}
\definecolor{listingorange}{HTML}{B84C16}
\definecolor{listinggray}{HTML}{737886}
\definecolor{listinglightgray}{HTML}{E8E9EF}
\definecolor{listingbackground}{HTML}{F7F7F9}

\definecolor{bsiocolor}{HTML}{B84C16}
\newcommand{\bsio}[1]{\textcolor{bsiocolor}{#1}}

\lstdefinestyle{codeblock}{
  backgroundcolor=\color{listingbackground},
  basicstyle=\ttfamily\scriptsize\color{listinggraphite},
  frame=Ltrb,
  framerule=0pt,
  rulecolor=\color{listingbackground},
  rulesepcolor=\color{listinggray},
  rulesep=1pt,
  breaklines=true,
  breakatwhitespace=false,
  columns=fullflexible,
  keepspaces=true,
  showstringspaces=false,
  xleftmargin=2pt,
  xrightmargin=2pt,
  aboveskip=4pt,
  belowskip=4pt,
  framesep=4pt,
  commentstyle=\color{listinggray}\itshape,
  stringstyle=\color{listingorange},
  keywordstyle=\color{listinggraphite},
  escapeinside={(*@}{@*)}
}

\lstdefinestyle{pytorchcode}{
  style=codeblock,
  language=Python,
  backgroundcolor=\color{listingbackground},
  rulecolor=\color{listingbackground},
  rulesepcolor=\color{listinggray},
  keywordstyle=\color{listingdarkpurple}\bfseries,
  commentstyle=\color{listinggray}\itshape,
  stringstyle=\color{listingorange},
  basicstyle=\tiny\ttfamily\color{listinggraphite}
}

\lstdefinelanguage{brainsurgerylang}{
  sensitive=false,
  comment=[l]{\#},
  commentstyle=\color{listinggray}\itshape,
  morekeywords={assert:,copy:,move:,cast:,cast_:,fill:,delete:,diff:,subtract_:,add_:,phlora:,phlora_:,assign:,concat:,scale_:,split:,reshape:,permute:,matmul:},
  keywordstyle=\color{listingdarkpurple}\bfseries,
  morekeywords=[2]{inputs:,output:,transforms:,from:,to:,target:,left:,right:,of:,is:,dtype:,mode:,value:,path:,format:,shard:,shape:,equal:,exists:,all:,not:,by:,rank:,target_a:,target_b:,left_alias:,right_alias:,sizes:,dim:,order:,from_a:,from_b:},
  keywordstyle=[2]\color{listingpurple}\bfseries,
  morestring=[b]",
  morestring=[b]',
  stringstyle=\color{listingorange},
  basicstyle=\ttfamily\scriptsize\color{listinggraphite},
  alsoletter={.\-:_},
  emph={assert:,copy:,move:,cast:,cast_:,fill:,delete:,diff:,subtract_:,add_:,phlora:,phlora_:,assign:,concat:,scale_:,split:,reshape:,permute:,matmul:},
  emphstyle=\color{listingdarkpurple}\bfseries,
  emph={[2]inputs:,output:,transforms:,from:,to:,target:,left:,right:,of:,is:,dtype:,mode:,value:,path:,format:,shard:,shape:,equal:,exists:,all:,not:,by:,rank:,target_a:,target_b:,left_alias:,right_alias:,sizes:,dim:,order:,from_a:,from_b:},
  emphstyle={[2]\color{listingpurple}\bfseries}
}

\lstdefinestyle{brainsurgeryyaml}{
  style=codeblock,
  language=brainsurgerylang,
  backgroundcolor=\color{listingbackground},
  rulecolor=\color{listingbackground},
  rulesepcolor=\color{listingdarkpurple},
  basicstyle=\tiny\ttfamily\color{listinggraphite}
}

\newcommand{\bscmd}[1]{\texttt{\textcolor{bsgraphite}{#1}}}
\newcommand{\bsfieldname}[1]{\texttt{\textcolor{bsgraphite}{#1}}}
\newcommand{\bslit}[1]{\textcolor{bsgraphite}{#1}}

\newcommand{\pyop}[1]{\textcolor{bsgraphite}{#1}}
\newcommand{\pykey}[1]{\textcolor{bsgraphite}{#1}}

\newcommand{\method}{\textsc{BrainSurgery}}

\usepackage{comment}
\usepackage{arydshln}
\usepackage{booktabs}
\usepackage{tabularx}

\title{\method{}:
Reproducible and Flexible Declarative Weight Manipulations for Model Editing and Upcycling}

\author{
    Gianluca Barmina\thanks{Equal contribution.}, Annemette Brok Pirchert\footnotemark[1], Andrea Blasi Núñez, \\ 
    \textbf{Lukas Galke Poech}, \textbf{Peter Schneider-Kamp} \\\\
    University of Southern Denmark \\
    \texttt{\{gbarmina,ampirchert,abln,galke,petersk\}@imada.sdu.dk}
}

\begin{document}
\maketitle

\begin{abstract}
Checkpoint transformations such as model merging, precision casting, and low-rank rewriting are commonly implemented through Python code or separate task-specific tools. We present \method{}, a tool that consolidates these operations into reusable declarative YAML plans over checkpoint tensors with built-in targeting, validation, diffing, and sharded export; a Web UI supports interactive inspection and plan construction. 
We compare \method{} with three specialized systems across 13 settings, 11 benchmarks, 8 models, 4 architectures, and model sizes from 1.7B to 32B parameters. Without use-case-specific framework customization, \method{} reproduces their model edits with comparable run cost, often lower memory usage, and close downstream performance. We further perform correctness, preservation, robustness, and scaling checks. Finally, a coding-agent study shows that agents can reliably generate and review \method{} plans, which are substantially shorter and more structurally consistent than equivalent Python implementations. These results support \method{} as a flexible alternative to maintaining separate tools for heterogeneous checkpoint-editing tasks.
\href{https://github.com/schneiderkamplab/brainsurgery}{\faGithub\ github.com/schneiderkamplab/brainsurgery}
\quad
\href{https://pypi.org/project/brainsurgery/}
{\faPython\ pypi.org/project/brainsurgery}
\end{abstract}

\section{Introduction}
\label{sec:introduction}

\begin{figure*}[!t]
    \centering
    \includegraphics[width=\linewidth]{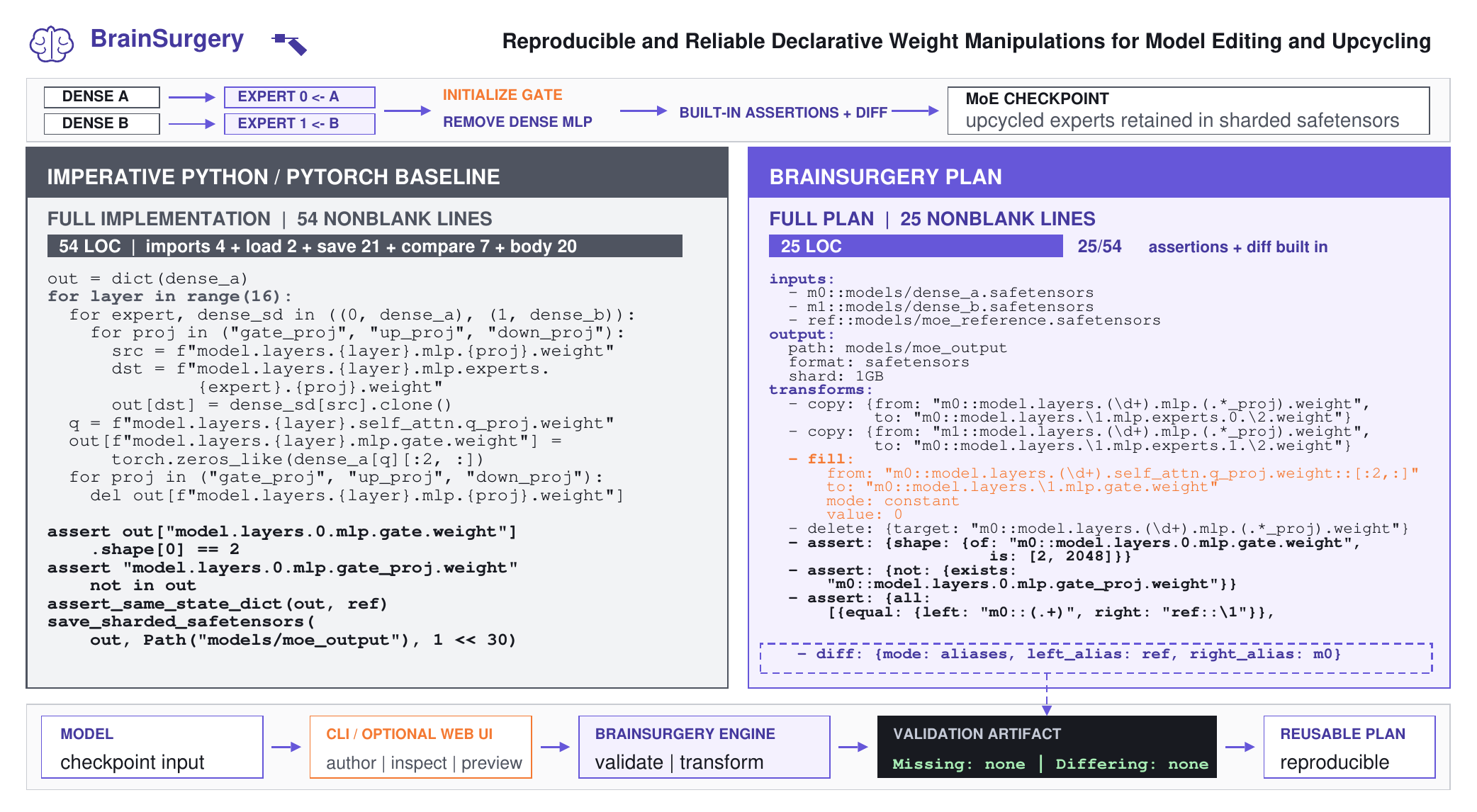}
    \caption{\textbf{\method{} workflow and dense-to-MoE example.} The same upcycling operation requires 54 nonblank lines in the imperative Python/PyTorch implementation and 25 in the \method{} plan. The plan maps two dense MLPs into separate experts, initializes the gate, removes the original dense MLP tensors, and performs built-in assertions and diffing. The lower panel summarizes the general workflow from model checkpoint and declarative plan to execution and a validation artifact.}
    \label{fig:infographics}
\end{figure*}

\begin{table*}[!t]
\centering
\scriptsize
\setlength{\tabcolsep}{3.2pt}
\renewcommand{\arraystretch}{1.10}
\begin{tabularx}{\textwidth}{@{}l*{8}{>{\centering\arraybackslash}X}@{}}
\toprule
\textbf{System} &
\shortstack{\textbf{Reusable}\\\textbf{specification}} &
\shortstack{\textbf{General tensor}\\\textbf{operations}} &
\shortstack{\textbf{Pattern-based}\\\textbf{targeting}} &
\shortstack{\textbf{Dtype}\\\textbf{control}} &
\shortstack{\textbf{Executable}\\\textbf{post-conditions}} &
\shortstack{\textbf{Checkpoint}\\\textbf{diff}} &
\shortstack{\textbf{Sharded}\\\textbf{export}} &
\shortstack{\textbf{Interactive}\\\textbf{UI}} \\
\midrule
Python/PyTorch scripts
& \faPython & \faPython & \faPython & \faPython & \faPython & \faPython & \faPython & \faPython \\
LLM-Pruner
& $\diamond$ & -- & $\diamond$ & -- & -- & -- & \faPython & $\diamond$ \\
MergeKit
& $\checkmark$ & $\diamond$ & $\diamond$ & $\checkmark$ & -- & -- & $\checkmark$ & \faUsers \\
Hugging Face PEFT
& $\diamond$ & -- & $\diamond$ & $\diamond$ & -- & -- & \faPython & -- \\
\texttt{torch-state-bridge}
& $\diamond$ & $\diamond$ & $\checkmark$ & -- & -- & $\checkmark$ & \faPython & -- \\
\midrule
\textbf{\method{}}
& $\checkmark$ & $\checkmark$ & $\checkmark$ & $\checkmark$ & $\checkmark$ & $\checkmark$ & $\checkmark$ & $\checkmark$ \\
\bottomrule
\end{tabularx}
\caption{\textbf{Integrated workflow coverage of the closest implementation alternatives.} $\checkmark$ denotes native general support, $\diamond$ denotes native but use-case-specific support, \faPython{} denotes either the user-written Python baseline or functionality that must be supplied by a surrounding Python workflow, \faUsers{} denotes a community-provided interface rather than an integrated first-party UI, and ``--'' denotes no integrated facility in the system's documented scope. The comparison concerns provided workflow capabilities (and not arbitrary extension code).
}
\label{tab:tool-comparison-compact}
\end{table*}

Post-hoc checkpoint manipulation is now a routine task in natural language processing model development: researchers and practitioners merge models and task vectors~\citep{ilharco2023editing,yadav2023tiesmerging,goddard2024mergekit}, integrate or extract low-rank adapters~\citep{hu2021lora}, prune or sparsify weights~\citep{cheng2024survey,he2023structured}, rename tensors across architectures, cast precision, shard checkpoints, and verify structural changes before deployment. The common implementation pattern is a Python script that loads a state dictionary, loops over tensor names, mutates values, checks post-conditions manually, and writes the resulting artifact. Such scripts are flexible, but they distribute the rewrite logic across I/O, control flow, mutation, validation, and persistence, making checkpoint edits difficult to audit, compare, and reproduce. The systems we mention in the paper partially solve these issues, providing tools for specialized model edits. However these are often tailored around one goal (e.g. pruning, merging), allowing only a few operations to be done.

We present \method{}, a system for declarative neural checkpoint editing (Figure~\ref{fig:infographics}). Users express weight-level transformations as YAML plans, execute them in batch or interactively, inspect checkpoint state through a Web UI, and validate local and end-to-end post-conditions with built-in assertions and diffs. \method{} operates directly on checkpoint files and aliases without model-class instantiation, providing one interface for transformations otherwise distributed across scripts and specialized tools. The transformation engine is architecture-agnostic: its operations apply to arbitrary named tensor mappings without model-family-specific rewrite code.

In sum, our contributions are:
\begin{itemize}[noitemsep,topsep=0pt,leftmargin=*]
    \item A unified declarative system for regex and structured tensor targeting, slicing, structural and arithmetic edits, dtype conversion, model merging, MoE upcycling, low-rank expert rewriting, and post-hoc low-rank adaptation~\citep{vasani2025phloradatafreeposthoclowrank}, with single-file and sharded checkpoint export and a Web UI.
    \item A comparison with independent systems on run costs and downstream performance, showing \method{} correctly implements such operations at comparable costs and without customization. 
    \item A study on coding-agents implementation and auditability, showing agents can reliably use \method{} and encode model edits in YAML plans that are shorter than equivalent generated Python.
\end{itemize}

\paragraph{Availability and demonstration.}
\method{} is released under the MIT license, is available on PyPI\footnote{\url{https://pypi.org/project/brainsurgery/}} and GitHub\footnote{\url{https://github.com/schneiderkamplab/brainsurgery}}. 
The video demonstration shows interactive checkpoint editing, alias-based diffing as reproducible plan execution\footnote{\url{https://shorturl.at/oCwhT}}.

\section{Related Systems}
\label{sec:related-systems}

Python code is the most common tool used for checkpoint editing. While it can express arbitrary checkpoint rewrites, it requires users to implement tensor selection, mutation, validation, and serialization themselves.
Python and PyTorch are in turn used to build systems that abstract custom operations in order to facilitate them.
The closest such systems are the three tools we compare with in Section~\ref{sec:tool-comparison}. LLM-Pruner~\citep{ma2023llmpruner} targets structural pruning with layer-, channel-, and block-wise methods and pruning-specific importance criteria. MergeKit~\citep{goddard2024mergekit} provides declarative recipes and specialized algorithms for model merging, layer slicing, and related construction; \method{} covers several of these transformations through general tensor operations. Neuroanatomy~\citep{ng2026llm} identify and evaluate layer-routing configurations. Some of these tools implement custom algorithms for e.g. layer-selection or routing configuration, which are outside \method{}'s scope and not evaluated. However, all of them apply algorithms' results to models through custom model edits, which is our focus and we carefully evaluate in Section~\ref{sec:tool-comparison}.
Unlike these use-case-specific systems, \method{} provides a more general and reusable interface for tensor targeting, ordered structural and arithmetic edits, dtype control, assertions, checkpoint diffing, and export. 
Other related systems are Orbax \citep{orbax} and \texttt{torch-state-bridge} \citep{torchstatebridge}, which cover narrower adjacent functionality: JAX-only checkpoint management and restoration, rule- and regex-based key remapping. Neither provides \method{}'s integrated combination of tensor-value transformations, checkpoint I/O and sharded export, and declarative CLI/UI workflow.

\section{\method{} Framework}
\label{sec:brainsurgery}

\method{} operates directly on checkpoint files without model classes instantiation. A plan specifies \bsfieldname{inputs} aliases, an ordered sequence of \bsfieldname{transforms}, and an optional \bsfieldname{output} artifact. Later transforms can refer to tensors created or modified earlier in the same plan, making the plan a reproducible record of the full checkpoint rewrite.

The transform library covers structural operations (\bscmd{copy}, \bscmd{move}, \bscmd{delete}, \bscmd{split}, \bscmd{concat}), shape and type operations (\bscmd{reshape}, \bscmd{permute}, \bscmd{cast}), arithmetic operations (\bscmd{add}, \bscmd{subtract}, \bscmd{scale}, \bscmd{matmul}), initialization (\bscmd{fill}), and low-rank factorization or reconstruction (\bscmd{phlora}). Tensor references can be written as regex patterns or structured path expressions, with optional slices. This gives users a single syntax for bulk targeting, local tensor extraction, dtype conversion, deletion, and advanced factorization workflows.

For reproducibility and scale, \method{} supports \texttt{safetensors} and PyTorch checkpoints, sharded output, batch execution, interactive execution, and multiple storage providers, including a memory-mapped arena provider for intermediate tensors and model copies. The Web UI exposes the same workflow interactively: users load aliases, configure \bsfieldname{transforms}, inspect model state, preview recent impact, compare aliases with checkpoint \bscmd{diff}, and export results, we defer Web UI details to Appendix~\ref{app:web-ui}.

\section{System Evaluation}

We frame the evaluation around two main questions: whether \method{} reproduces edits from existing systems while achieving comparable downstream performance, a practically relevant outcome for end users, and whether coding agents can use it effectively, given their increasing use in coding workflows. We complement these evaluations with finer-grained validation checks of transformation correctness, preservation, robustness, scaling and low-rank and MoEs workflows. We include additional checks across ten checkpoints in Appendix~\ref{app:behavioral-analysis}.

\subsection{System Comparison}
\label{sec:tool-comparison}

\begin{table*}[t]
\centering
\scriptsize
\setlength{\tabcolsep}{5pt}
\renewcommand{\arraystretch}{1.15}
\begin{tabular}{llrccccc}
\toprule
\textbf{Reference tool} &
\textbf{Mode} &
\textbf{Cases} &
\shortstack{\textbf{Throughput} \textbf{ratio}} &
\shortstack{\textbf{Peak PSS} \textbf{ratio}} &
\shortstack{\textbf{Avg PSS} \textbf{ratio}} &
\shortstack{\textbf{Logical I/O} \textbf{ratio}} &
\shortstack{\textbf{Physical I/O} \textbf{ratio}} \\
\midrule

LLM-Pruner
& arena
& 3
& $0.37_{\scriptscriptstyle [0.35\text{--}0.55]}$
& $0.50_{\scriptscriptstyle [0.50\text{--}0.56]}$
& $0.59_{\scriptscriptstyle [0.58\text{--}0.91]}$
& $0.64_{\scriptscriptstyle [0.63\text{--}0.64]}$
& $2.96_{\scriptscriptstyle [2.31\text{--}2.96]}$ \\

LLM-Pruner
& in-memory
& 3
& $0.83_{\scriptscriptstyle [0.78\text{--}1.37]}$
& $0.32_{\scriptscriptstyle [0.32\text{--}0.40]}$
& $0.34_{\scriptscriptstyle [0.34\text{--}0.35]}$
& $0.64_{\scriptscriptstyle [0.63\text{--}0.64]}$
& $1.00_{\scriptscriptstyle [1.00\text{--}1.00]}$ \\

MergeKit
& arena
& 9
& $0.69_{\scriptscriptstyle [0.52\text{--}0.89]}$
& $0.75_{\scriptscriptstyle [0.53\text{--}1.12]}$
& $1.04_{\scriptscriptstyle [0.57\text{--}1.53]}$
& $1.00_{\scriptscriptstyle [1.00\text{--}1.00]}$
& $3.01_{\scriptscriptstyle [2.00\text{--}4.20]}$ \\

MergeKit
& in-memory
& 9
& $0.97_{\scriptscriptstyle [0.61\text{--}2.33]}$
& $0.53_{\scriptscriptstyle [0.40\text{--}1.00]}$
& $0.86_{\scriptscriptstyle [0.38\text{--}1.31]}$
& $1.00_{\scriptscriptstyle [1.00\text{--}1.00]}$
& $1.00_{\scriptscriptstyle [1.00\text{--}1.00]}$ \\

LLM Neuroanatomy
& arena
& 1
& 0.40
& 1.08
& 1.85
& 0.98
& 2.00 \\

LLM Neuroanatomy
& in-memory
& 1
& 0.97
& 1.05
& 1.25
& 0.98
& 1.00 \\

\bottomrule
\end{tabular}
\caption{
Run-cost ratios (\method{} divided by the reference tool), aggregated by reference tool and execution mode across all evaluated cases. Per-case median is used.
Higher throughput ratios favor \method{} when above 1, while lower PSS and logical/physical-I/O ratios favor it when below 1. Bracketed values report the range across cases, not computable for LLM Neuroanatomy as single-case.
}
\label{tab:run-cost-aggregate}
\end{table*}

\begin{table*}[t]
\centering
\small
\setlength{\tabcolsep}{5pt}
\renewcommand{\arraystretch}{1.15}
\resizebox{\textwidth}{!}{%
\begin{tabular}{llrrr}
\toprule
\textbf{Reference tool} &
\textbf{Benchmarks} &
\shortstack{\textbf{Total}\\\textbf{comparisons}} &
\shortstack{\textbf{Statistically}\\\textbf{indistinguishable}} &
\shortstack{\textbf{Avg $|\Delta|$}} \\
\midrule

LLM-Pruner
& HellaSwag, BoolQ, OpenBookQA, ARC-E/C, WinoGrande, PIQA
& 36
& 36
& $0.00_{\scriptscriptstyle [0.00\text{--}0.00]}$ \\

MergeKit
& HellaSwag, MMLU
& 27
& 21
& $0.23_{\scriptscriptstyle [0.00\text{--}1.36]}$ \\

LLM Neuroanatomy
& BBH, GSM8K, IFEval, MBPP
& 11
& 11
& $0.00_{\scriptscriptstyle [0.00\text{--}0.00]}$ \\

\bottomrule
\end{tabular}%
}
\caption{
Downstream-evaluation parity between \method{} and each specialized reference tool.
Metric-level comparisons are statistically indistinguishable if the confidence interval includes zero, or the scores match exactly.
The final column reports absolute arithmetic mean of differences (percentage points) across all comparisons. Bracketed values report the range across cases.
}
\label{tab:downstream-parity}
\end{table*}

We compare \method{} with three specialized third-party implementations of common and useful model edits expressible through \method{} plans, evaluating run cost (throughput, memory usage, I/O operations) and downstream performance. Overall, the comparison spans 13 settings, 11 benchmarks, 8 models, 4 architectures, and model sizes from 1.7B to 32B parameters. The results show that \method{}, particularly the in-memory version, remains comparable in run cost while preserving downstream capabilities, without requiring use-case-specific customization of \method{}'s code.

We consider three existing specialized implementations: (i) LLM Neuroanatomy, based on insights from \citet{ng2026llm} and an independent reproduction\footnote{\url{https://github.com/alainnothere/llm-circuit-finder}}, which show that duplicating selected layers can improve downstream performance; (ii) MergeKit \citep{goddard2024mergekit}, a framework for model merging that exploits transfer learning without additional training; and (iii) LLM-Pruner \citep{ma2023llmpruner}, which performs structural pruning to improve efficiency while preserving model functionality. For each system, we use models, tasks, and settings explicitly documented in the corresponding repository.

For LLM Neuroanatomy, we closely follow the Qwen-32B-Coder~\citep{hui2024qwen25codertechnicalreport} example by duplicating layers 7--9 once. For MergeKit, we evaluate three base/instruct pairs, SmolLM2-1.7B~\citep{allal2025smollm2}, Qwen2.5-7B~\citep{qwen2025qwen25technicalreport}, and OLMo-2-32B~\citep{olmo20242olmo2furious}, under three transformations: passthrough layer stacking using the first 60\% of layers followed by the suffix starting at 50\%, linear merging ($0.4 \times \text{base} + 0.6 \times \text{instruct}$), and task arithmetic ($\text{base} + 0.75 \times (\text{instruct} - \text{base})$). For LLM-Pruner, we evaluate LLaMA-7B~\citep{touvron2023llama} under layer-wise pruning to layers 0--23, block-L2 pruning with 25\% local pruning of attention heads and MLP channels across layers 4--29, and block-Taylor pruning with the same pruning rate and layer range using Taylor-importance decisions.

\paragraph{Experimental setup.}
For run cost evaluation, we measure throughput in input GiB/s, peak and average proportional set size (PSS), logical model I/O, and physical I/O, all normalized by input size where applicable. PSS measures a process's effective memory usage by counting shared memory proportionally to how many processes use it. Logical I/O measures tensor payloads at the model loader/writer boundary, while physical I/O measures storage-layer read and write bytes. Each configuration is run five times, with parallelism fixed at 50 compute threads whenever supported by the reference implementation. 

We evaluate downstream performance with \texttt{lm-evaluation-harness} \citep{eval-harness}, following the benchmarks and settings reported in the published repositories for LLM-Pruner and LLM Neuroanatomy. MergeKit does not specify downstream evaluation benchmarks for the use cases covered, so we use two widely adopted benchmarks, HellaSwag \citep{zellers-etal-2019-hellaswag} and MMLU \citep{hendrycks2021mmlu}. Comparisons are done across benchmarks and metrics. Each comparison is between a benchmark's metric score measured for the reference system and the same score measured for \method{}. Arena and in-memory executions are collapsed into a single comparison, as both produce identical outputs.

\paragraph{Preserving performance, increasing flexibility.}
We report downstream results in Table~\ref{tab:downstream-parity}. For each benchmark and metric, we compute the delta between the reference system's score and \method{}'s score and a 95\% confidence interval via a paired bootstrap (10{,}000 resamples, seed 0) over matched per-example predictions. A comparison is statistically indistinguishable when this interval includes zero. Under this criterion, all 36 LLM-Pruner and 11 LLM Neuroanatomy comparisons are statistically indistinguishable, while 21 of 27 MergeKit comparisons are indistinguishable, with an average absolute difference of 0.23 percentage points (arithmetic mean across all 27 comparisons).
For run cost, in-memory execution achieves comparable throughput and is more memory-efficient in two of the three use cases, reducing peak and average PSS by up to 68\% and 66\%, respectively, while matching or improving logical and physical I/O. Arena execution is generally slower and incurs higher physical I/O, reflecting its use of memory-mapped storage for intermediate tensors and model copies, which trades memory pressure for additional storage traffic. Overall \method{}, particularly its in-memory execution, achieves comparable run cost and preserves downstream capabilities across three distinct editing use cases within a single framework and, importantly, without any use-case-specific customization of \method{}'s code.
We provide further details and full results in Appendix~\ref{app:full-res-system-comparison}.

\subsection{Coding-Agent Usability and Auditability}
\label{sec:agent-evaluation}

\begin{figure}[!t]
    \centering
    \includegraphics[width=\linewidth]{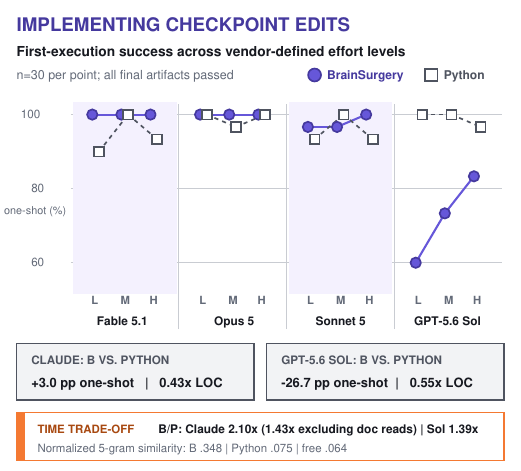}
    \caption{First-execution success of agents' generated Python and \method{} plans over 30 fresh sessions at one effort level. 
    }
    \label{fig:agent-implementation}
\end{figure}

Agents are increasingly used for coding tasks, therefore we evaluate how easily and reliably they can use \method{} for model edits measuring first-execution reliability, implementation size, similarity across independent solutions, and cold-start time and cost.
The experiment compares \method{} plans (B), Python/PyTorch implementations (P), and agent-selected tools from a related-tool allow-list (F). 
It comprises $1{,}080$ fresh sessions: five tasks, three target models, four agents, three vendor-defined effort levels, two repeats, and three implementation conditions. After each session terminates, a deterministic Python script checks the final checkpoint's loadability, tensor names, shapes, dtypes, values, and required sharding. All generated artifacts pass; first-execution success therefore measures whether the initial implementation works before any repair.

\paragraph{Usability.}
Figure~\ref{fig:agent-implementation} shows that coding agents can use \method{} effectively, with all $360$ BrainSurgery sessions eventually producing a correct artifact and high first-execution success overall. The results are agent-dependent: Claude succeeds on the first execution in 268/270 B sessions (99.3\%), slightly above P sessions at 260/270 (96.3\%), while GPT-5.6 Sol succeeds in 65/90 B sessions (72.2\%) compared with 89/90 (98.9\%) for P, corresponding to paired differences of $+3.0$ points (95\% CI $[+1.1,+5.2]$) and $-26.7$ points (95\% CI $[-40.0,-14.4]$), respectively.

We compare counts of non-blank, non-comment code lines in the \method{} plans and executable implementations, both generated by agents, by calculating
 $\operatorname{median}(\mathrm{LOC}_{B})/
\operatorname{median}(\mathrm{LOC}_{P})$.
The ratios are 0.43 for Claude and 0.55 for Sol, hence generated \method{} plans are 57\% and 45\% shorter, respectively.

We compare code similarity among solutions generated in different sessions for the same task, target model, and condition. After removing blank, comment-only lines and normalizing filesystem paths, we compute the Jaccard similarity between token 5-gram sets. Across 4,140 eligible pairs per condition, median similarity is 0.348 for \method{}, 0.075 for Python/PyTorch, and 0.064 for free-choice tools, indicating that \method{} plans converge on a substantially more consistent implementation form.

\paragraph{Artifact auditability.}
In a separate review experiment, fresh agent sessions evaluate either a correct artifact or one containing a single injected defect. Agents review \method{} plans with high accuracy, reaching 94.2\% balanced accuracy (the mean of defect-rejection and correct-acceptance rates) and rejecting all defective artifacts; this is comparable to Python/PyTorch (95.0\%) and free-choice tools (95.3\%), with no significant B--P difference ($-1.0$ points, 95\% CI $[-2.9,+0.7]$). All errors are false rejections of correct artifacts. Full results appear in Table~\ref{tab:agent-auditability} of Appendix~\ref{app:coding-agent-evaluation}.

\subsection{Validation Checks}
\label{sec:validation}

\paragraph{Transformation correctness.}
For correct execution, ten transformations use hand-verifiable fixtures, and three real-checkpoint cases apply an explicit multiply-by-one operation to one tensor in GPT-2 124M~\citep{radford2019language}, OLMo-1B-0724-hf~\citep{Groeneveld2023OLMo}, and Pythia-1B~\citep{biderman2023pythiasuiteanalyzinglarge}. Using PyTorch, we separately implement a verifier that uses neither \method{} nor its outputs as expectations and checks tensor names, shapes, dtypes, bytes, source files, and tensors outside the declared write set. All 679 output-tensor comparisons match, all 664 out-of-scope comparisons remain byte-identical, and every source remains unchanged.

\paragraph{Arithmetic precision.}
We apply multiplication by $0.5$ followed by multiplication by $2$ to four Pythia checkpoints to distinguish finite-precision effects from transformation errors. \method{} and direct PyTorch produce identical restored checkpoints and behavior in every condition. Native FP16 loses information, whereas FP32 restores every tensor and measured behavior exactly; the FP16 differences therefore arise from finite-precision arithmetic rather than a \method{}-specific error. Complete results appear in Appendix~\ref{app:behavioral-analysis}.

\paragraph{Scaling and resource use.}
We measure one checkpoint-wide rewrite on ten checkpoints using direct Python/PyTorch and \method{}'s in-memory and arena providers. All 150 measured outputs pass an author-written output oracle that verifies the complete tensor set, values, dtypes, shard index, shard-size constraint, and unchanged source files. At Pythia 12B, \method{}'s in-memory provider reduces peak RSS by 48.9\% (22.88 versus 44.77~GiB), while taking $1.59\times$ as long as direct Python/PyTorch (24.59 versus 15.50~s). Appendix~\ref{app:scaling} provides the complete protocol and per-checkpoint results.

\paragraph{Failure behavior.}
We test \method{} robustness using 18 cases: 15 before-save failures, 3 faults injected after saving begins. 
The before-save failures occur during execution of the plan and cover malformed plans, invalid transformations, invalid tensor matches and references, missing or damaged inputs, failed assertions, and a pre-existing destination. The remaining 3 simulate: a save failure after one shard is written to a fresh destination, process interruption at the same point, and a save failure while writing to a pre-existing destination.
Corrupted plans are not executed and source checkpoints remain unchanged in all cases. Thus, the tested errors are detected without damaging their inputs, but an interrupted save can leave a partial checkpoint. Complete results appear in Appendices~\ref{app:correctness-preservation} and~\ref{app:robustness}.

\paragraph{MoE and low-rank workflow.}
The \method{} framework provides reusable plans for dense-to-expert MoE upcycling~\citep{shi2025flexolmoopenlanguagemodels}, PHLoRA factorization~\citep{vasani2025phloradatafreeposthoclowrank,pirchert2026flexmoreflexiblemixturerankheterogeneous}, and in-place low-rank expert rewriting. Figure~\ref{fig:infographics} presents a comparison of the MoE plan and its imperative baseline.  For each workflow, a separate author-written imperative Python baseline script performs the same checkpoint rewrite. The complete \method{} and baseline outputs match in tensor names, shapes, dtypes, and values, and all post-conditions pass. We include details of tests on OLMo-1B-0724-hf~\citep{Groeneveld2023OLMo} in Appendix~\ref{app:examples-cases}.

\section{Discussion}
\label{sec:discussion}

Our results support \method{} as a flexible framework for inspectable checkpoint editing. It implements bulk edits, merging, MoE upcycling, low-rank rewriting, PHLoRA factorization, validation, and sharded export. Our evaluations in Section~\ref{sec:tool-comparison} covered multiple types of pruning, merging, and layer duplication by three different third-party systems. We showed that \method{} correctly reproduced the results with close downstream scores, comparable run costs, and more importantly, without extending the framework. Coding agents are able to use \method: one-shot correctness of generated YAML plans is high overall, plans are substantially shorter than equivalent generated Python, and more consistent.

\section{Conclusion}
\label{sec:conclusion}

\method{} provides a single framework for checkpoint edits otherwise handled by separate specialized systems. Without use-case-specific customization, it reproduces edits from three third-party tools while achieving close downstream performance and comparable run cost, with in-memory execution often reducing memory usage. Coding agents can also use \method{} effectively: all generated artifacts are eventually correct, plans are substantially shorter and more structurally consistent than equivalent Python implementations. Overall, \method{} combines flexibility, efficiency, and compact declarative model editing within one reusable framework.

\section*{Limitations}
\label{sec:limitations}

The system comparison is limited to the systems and cases considered in our evaluation. The results therefore should not be generalized to other tools, transformations, models, or experimental settings. \method{} does not implement custom algorithms to determine which model edits to apply (e.g. pruning methods from LLM Pruner), but provides a more flexible and reproducible way for applying them. 

A human-participant study was not feasible within the available resources because of its cost. Instead, we collected informal internal feedback from members of the research group. This feedback does not replace a controlled user study. We complemented it with the coding-agent study presented in Section~\ref{sec:agent-evaluation}, motivated by the increasing use of agents for coding tasks. The results of this study characterize the evaluated agents and should not be interpreted as evidence of human usability.

The set of transforms we implement is based on operations we found to be common in model editing, as reflected in Section~\ref{sec:tool-comparison}. However, additional useful operations may exist and could be added. For this reason, \method{} is under continuous development and improvement.

\section*{Ethics and Broader Impact Statement} 

Due to resource constraints, we did not conduct a human-participant study but assessed coding agents' capabilities; this choice is also motivated by the increasing use of agents in coding workflows. However, \method{} is not designed specifically or exclusively for agents: it is intended to be used by human researchers and practitioners as well, and we encourage its use and evaluation by human users.

\method{} is intended to make checkpoint editing more reproducible and inspectable, but it does not determine whether a model modification is authorized, safe, or appropriate. Easier checkpoint manipulation may also lower the barrier to unintended or harmful modifications.

\section*{Acknowledgements}
The research was supported in part by the Danish Foundation Models project, funded by the Danish government.
This research was further supported in part by the MIST project, funded by the Novo Nordisk Foundation under grant reference number NNF25OC0103204.
Part of the computation for this project was performed on the UCloud interactive HPC system managed by the eScience Center at the University of Southern Denmark.

\bibliography{custom}

\appendix

\section{Web UI}
\label{app:web-ui}

\begin{figure*}[!h]
    \centering
    \includegraphics[width=\linewidth]{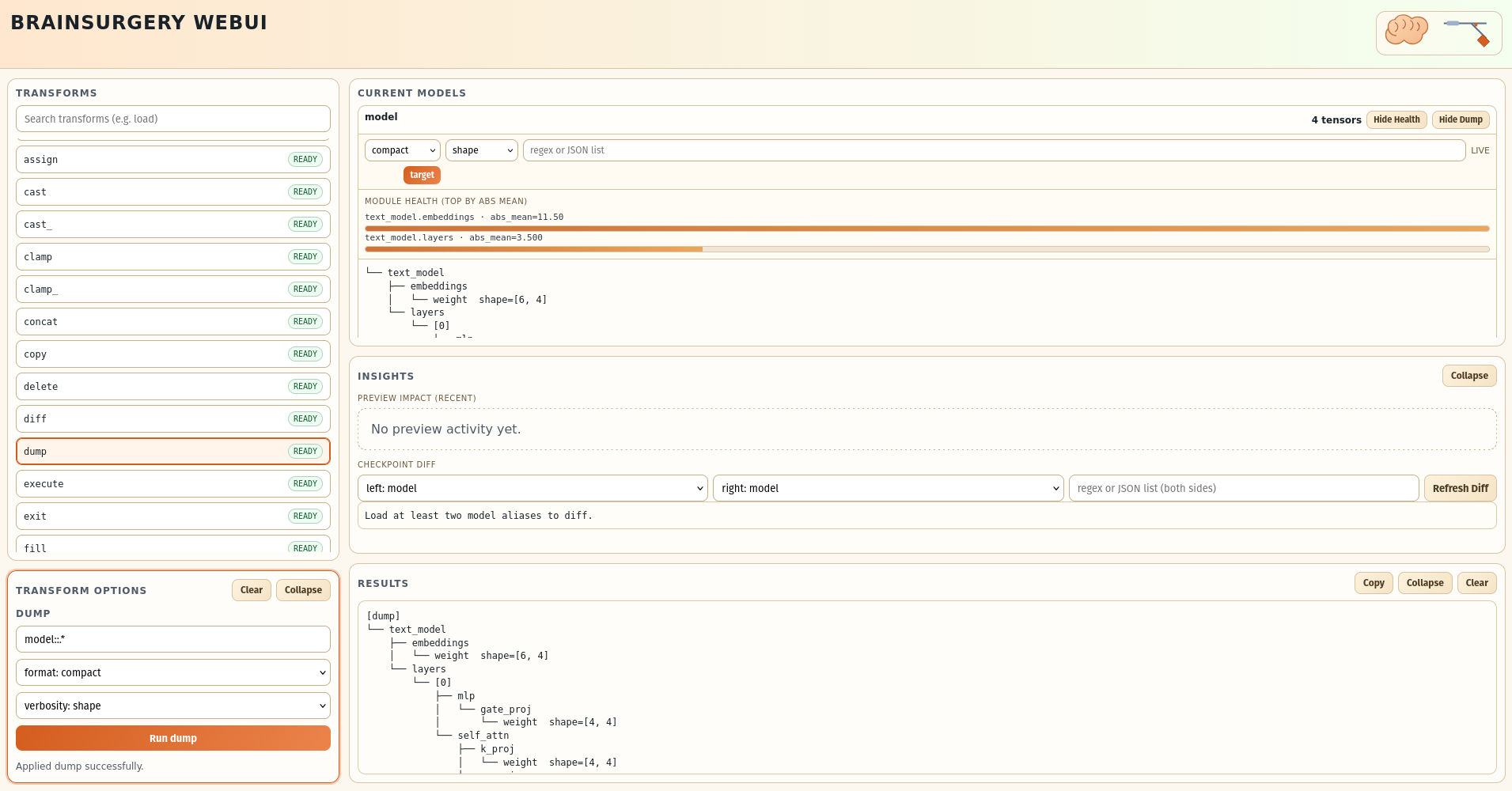}
    \caption{The \method{} Web UI for interactive checkpoint inspection, transform execution, previewing edit effects, and checkpoint export.}
    \label{fig:webui-dump}
\end{figure*}

\begin{figure*}[!h]
    \centering
    \includegraphics[width=\linewidth]{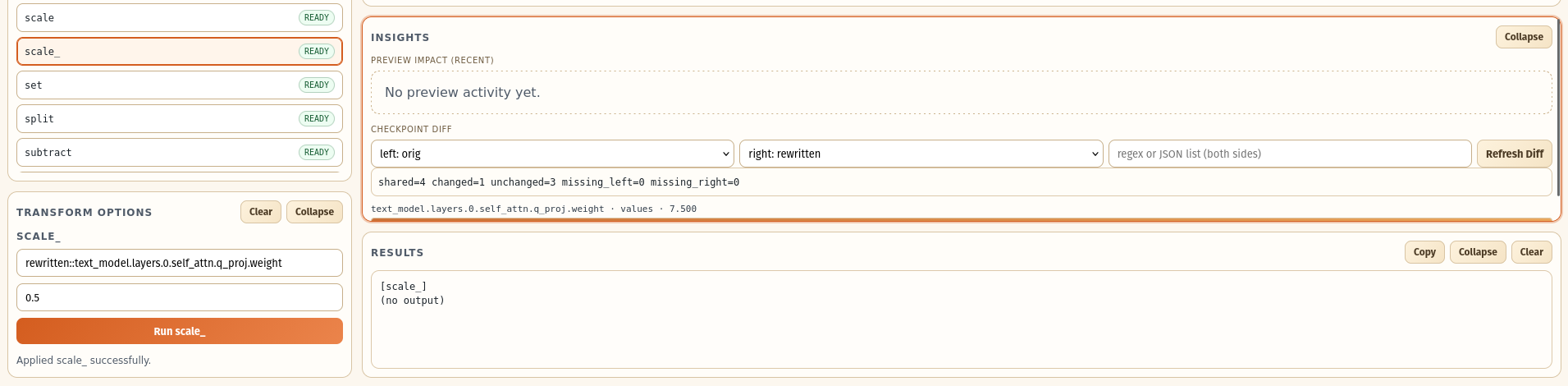}
    \caption{\method{} Web UI figure showing zoom-in on \bscmd{diff} between the original model and the rewritten model after applying \bscmd{scale\_}.}
    \label{fig:webui-zoom}
\end{figure*}

In the demo video, we present the interactive Web UI (shown in Figure~\ref{fig:webui-dump}--\ref{fig:webui-zoom}) version of workflows such as: load checkpoint aliases, apply a transform, inspect the result and validate with diff for OLMo-1B-0724-hf.

\section{Examples \& Case Studies}
\label{app:examples-cases}

This Appendix presents \method{} through a compact progression of examples (Figure~\ref{fig:twocol-slicing-appendix}--\ref{fig:validation-diff-snippet-appendix}),  3 case studies (OLMo-1B-0724-hf): full dense-to-expert MoE Figure~\ref{fig:full-moe-workflow}, PHLoRA factorization~\ref{fig:full-phlora-workflow} and full in-place low-rank expert rewrite with validation Figure~\ref{fig:full-low-rank-workflow}. We recognize, that there is not only one imperative way to express the corresponding rewrite in Python, regular expressions (\texttt{re}) and PyTorch (\texttt{torch}).

\begin{figure}[!h]
\centering
\begin{minipage}[t]{0.98\columnwidth}
\textbf{Imperative Python baseline}
\begin{lstlisting}[style=pytorchcode]
w = sd["model.layers.0.self_attn.q_proj.weight"]
sd["tmp"] = w[:(*@\bslit{128}@*), :(*@\bslit{128}@*)].(*@\pykey{clone}@*)()
\end{lstlisting}
\end{minipage}

\vspace{3pt}

\begin{minipage}[t]{0.98\columnwidth}
\textbf{\method{} transform fragment}
\begin{lstlisting}[style=brainsurgeryyaml]
- copy: from: ".*\.0\..*\.self_attn.q_proj.*::[:128, :128]", to: "tmp"

\end{lstlisting}
\end{minipage}
\caption{\textbf{Example tensor slicing}. Both sides copy the same \texttt{[:128, :128]} block into the same destination tensor slot.}
\label{fig:twocol-slicing-appendix}
\end{figure}

\begin{figure}[!h]
\centering
\begin{minipage}[t]{0.98\columnwidth}
\textbf{Imperative Python/PyTorch baseline}
\begin{lstlisting}[style=pytorchcode]
import torch

(*@\pyop{assert}@*) "layers.0.gate.weight" in out
(*@\pyop{assert}@*) out["layers.0.gate.weight"](*@\pykey{.shape}@*) == ((*@\bslit{2}@*), (*@\bslit{2048}@*))
(*@\pyop{assert}@*) (*@\pykey{torch.equal}@*)(
    src["layers.0.gate.weight"][:(*@\bslit{16}@*), :(*@\bslit{16}@*)],
    out["layers.0.gate.weight"][:(*@\bslit{16}@*), :(*@\bslit{16}@*)],
)
(*@\pyop{assert}@*) "layers.0.gate.bias" not in out
\end{lstlisting}
\end{minipage}

\vspace{3pt}

\begin{minipage}[t]{0.98\columnwidth}
\textbf{\method{} transform fragment}
\begin{lstlisting}[style=brainsurgeryyaml]
- assert: exists: "out::layers.0.gate.weight"
- assert: shape: { of: "out::layers.0.gate.weight", is: [(*@\bslit{2}@*), (*@\bslit{2048}@*)] }
- assert: equal:
    left: "src::layers.0.gate.weight::[:16, :16]"
    right: "out::layers.0.gate.weight::[:16, :16]"
- assert: not: { exists: "out::layers.0.gate.bias" }
\end{lstlisting}
\end{minipage}
\caption{\textbf{Example validation as executable invariants}. Both sides check the same existence, shape, equality, and deletion post-conditions.}
\label{fig:twocol-verification-appendix}
\end{figure}

\begin{figure}[!h]
\centering
\begin{minipage}[t]{0.98\columnwidth}
\textbf{Imperative Python/Re baseline}
\begin{lstlisting}[style=pytorchcode]
import re
import torch

sd = torch.load("(*@\bsio{models/input.pt}@*)")
pattern = re.compile(r".*self_attn\..*_proj\.weight")
for name, tensor in sd.items():
    if pattern.fullmatch(name):
        sd[name] = tensor * (*@\bslit{0.5}@*)
torch.save(sd, "(*@\bsio{models/output.pt}@*)")
\end{lstlisting}
\end{minipage}

\vspace{3pt}

\begin{minipage}[t]{0.98\columnwidth}
\textbf{\method{} transform}
\begin{lstlisting}[style=brainsurgeryyaml]
inputs: [ (*@\bsio{models/input.pt}@*) ]
scale_: target: ".*self_attn\..*_proj\.weight", by: (*@\bslit{0.5}@*)
output: (*@\bsio{models/output.pt}@*)
\end{lstlisting}
\end{minipage}
\caption{\textbf{Bulk tensor targeting}. The imperative baseline loops over matching checkpoint names; the \method{} fragment states the same regex target family and scale operation declaratively.}
\label{fig:twocol-bulk-targeting}
\end{figure}

\begin{figure}[!h]
\centering
\begin{minipage}[t]{0.485\textwidth}
\textbf{Imperative Python/PyTorch baseline}
\begin{lstlisting}[style=pytorchcode]
import torch
from safetensors.torch import load_file

yaml_out = load_file("(*@\bsio{models/example\_yaml\_output}@*)")
ref_out = load_file("(*@\bsio{models/example\_reference\_output}@*)")

missing_on_left = sorted(set(yaml_out) - set(ref_out))
missing_on_right = sorted(set(ref_out) - set(yaml_out))
differing = []

for name in sorted(set(yaml_out) & set(ref_out)):
    if yaml_out[name](*@\pykey{.shape}@*) != ref_out[name](*@\pykey{.shape}@*):
        differing.append((name, f"shape {tuple(ref_out[name].shape)} != {tuple(yaml_out[name].shape)}"))
    elif yaml_out[name](*@\pykey{.dtype}@*) != ref_out[name](*@\pykey{.dtype}@*):
        differing.append((name, f"dtype {ref_out[name].dtype} != {yaml_out[name].dtype}"))
    elif yaml_out[name](*@\pykey{.device}@*) != ref_out[name](*@\pykey{.device}@*):
        differing.append((name, f"device {ref_out[name].device} != {yaml_out[name].device}"))
    elif not (*@\pykey{torch.equal}@*)(yaml_out[name], ref_out[name]):
        differing.append((name, "values differ"))

print("Diff: ref::.* <-> yaml::.*")
print("Missing on left:")
if not missing_on_left:
    print("  (none)")
else:
    for name in missing_on_left:
        print(f"  - {name}")
print("Missing on right:")
if not missing_on_right:
    print("  (none)")
else:
    for name in missing_on_right:
        print(f"  - {name}")
print("Differing:")
if not differing:
    print("  (none)")
else:
    for name, reason in differing:
        print(f"  - {name}: {reason}")
if not missing_on_left and not missing_on_right and not differing:
    print("No differences found.")
\end{lstlisting}
\end{minipage}
\hfill
\begin{minipage}[t]{0.485\textwidth}
\textbf{\method{} validation artifact}
\begin{lstlisting}[style=brainsurgeryyaml]
inputs:
  - (*@\bslit{yaml}@*)::(*@\bsio{models/example\_yaml\_output}@*)
  - (*@\bslit{ref}@*)::(*@\bsio{models/example\_reference\_output}@*)

transforms:
  - diff: { mode: aliases, left_alias: (*@\bslit{ref}@*), right_alias: (*@\bslit{yaml}@*) }
\end{lstlisting}
\end{minipage}
\caption{\textbf{Validation with} \bscmd{diff}. Local invariants use \bscmd{assert}; end-to-end agreement with an independent reference is checked by diffing the reference alias against the \method{} output.}
\label{fig:validation-diff-snippet-appendix}
\end{figure}

\begin{figure*}[!h]
\centering
\begin{minipage}[t]{0.485\textwidth}
\textbf{Imperative Python/PyTorch baseline}
\begin{lstlisting}[style=pytorchcode]
from pathlib import Path
import json
import torch
from safetensors.torch import load_file, save_file

def load_checkpoint(path):
    return load_file(str(path)) if path.suffix == ".safetensors" else torch.load(path, weights_only=True)

def save_sharded_safetensors(sd, out_dir, max_bytes):
    out_dir.mkdir(parents=True, exist_ok=True)
    shards, cur, cur_size = [], {}, (*@\bslit{0}@*)
    for name, tensor in sd.items():
        size = tensor.numel() * tensor.element_size()
        if cur and cur_size + size > max_bytes:
            shards.append(cur)
            cur, cur_size = {}, (*@\bslit{0}@*)
        cur[name] = tensor
        cur_size += size
    if cur:
        shards.append(cur)
    weight_map = {}
    for idx, shard in enumerate(shards, start=(*@\bslit{1}@*)):
        shard_name = f"model-{idx:05d}-of-{len(shards):05d}.safetensors"
        save_file(shard, str(out_dir / shard_name))
        for name in shard:
            weight_map[name] = shard_name
    (out_dir / "model.safetensors.index.json").write_text(
        json.dumps({"weight_map": weight_map}), encoding="utf-8"
    )

def assert_same_state_dict(left, right):
    missing_l = sorted(set(right) - set(left))
    missing_r = sorted(set(left) - set(right))
    differing = [k for k in set(left) & set(right)
                 if left[k](*@\pykey{.shape}@*) != right[k](*@\pykey{.shape}@*)
                 or not (*@\pykey{torch.equal}@*)(left[k], right[k])]
    (*@\pyop{assert}@*) missing_l == missing_r == differing == []

dense_a = load_checkpoint(Path("(*@\bsio{models/dense\_a.safetensors}@*)"))
dense_b = load_checkpoint(Path("(*@\bsio{models/dense\_b.safetensors}@*)"))
ref = load_checkpoint(Path("(*@\bsio{models/moe\_reference.safetensors}@*)"))
out = dict(dense_a)

for layer in range((*@\bslit{16}@*)):
    for expert, dense_sd in (((*@\bslit{0}@*), dense_a), ((*@\bslit{1}@*), dense_b)):
        for proj in ("gate_proj", "up_proj", "down_proj"):
            src = f"model.layers.{layer}.mlp.{proj}.weight"
            dst = f"model.layers.{layer}.mlp.experts.{expert}.{proj}.weight"
            out[dst] = dense_sd[src].(*@\pykey{clone}@*)()
    q = f"model.layers.{layer}.self_attn.q_proj.weight"
    out[f"model.layers.{layer}.mlp.gate.weight"] = (*@\pykey{torch.zeros\_like}@*)(
        dense_a[q][:(*@\bslit{2}@*), :]
    )
    for proj in ("gate_proj", "up_proj", "down_proj"):
        (*@\pyop{del}@*) out[f"model.layers.{layer}.mlp.{proj}.weight"]

(*@\pyop{assert}@*) out["model.layers.0.mlp.gate.weight"](*@\pykey{.shape}@*)[(*@\bslit{0}@*)] == (*@\bslit{2}@*)
(*@\pyop{assert}@*) "model.layers.0.mlp.gate_proj.weight" not in out
assert_same_state_dict(out, ref)
save_sharded_safetensors(out, Path("(*@\bsio{models/moe\_output}@*)"), (*@\bslit{1}@*) << (*@\bslit{30}@*))
\end{lstlisting}
\end{minipage}
\hfill
\begin{minipage}[t]{0.485\textwidth}
\textbf{\method{} plan}
\begin{lstlisting}[style=brainsurgeryyaml]
inputs:
  - m0::(*@\bsio{models/dense\_a.safetensors}@*)
  - m1::(*@\bsio{models/dense\_b.safetensors}@*)
  - ref::(*@\bsio{models/moe\_reference.safetensors}@*)
output:
  path: (*@\bsio{models/moe\_output}@*)
  format: safetensors
  shard: (*@\bslit{1GB}@*)
transforms:
  - copy: { from: "m0::model.layers\.(\d+)\.mlp\.(.*_proj)\.weight", to: "m0::model.layers.\1.mlp.experts.0.\2.weight" }
  - copy: { from: "m1::model.layers\.(\d+)\.mlp\.(.*_proj)\.weight", to: "m0::model.layers.\1.mlp.experts.1.\2.weight" }
  - fill:
      from: "m0::model.layers\.(\d+)\.self_attn\.q_proj\.weight::[:2,:]"
      to: "m0::model.layers.\1.mlp.gate.weight"
      mode: constant
      value: (*@\bslit{0}@*)
  - delete: { target: "m0::model.layers\.(\d+)\.mlp\.(.*_proj)\.weight" }
  - assert:
      shape: { of: "m0::model.layers.0.mlp.gate.weight", is: [(*@\bslit{2}@*), (*@\bslit{2048}@*)] }
  - assert:
      not: { exists: "m0::model.layers.0.mlp.gate_proj.weight" }
  - assert:
      all:
        - equal: { left: "m0::(.+)", right: "ref::\1" }
        - equal: { left: "ref::(.+)", right: "m0::\1" }
\end{lstlisting}
\end{minipage}
\caption{\textbf{Full dense-to-expert MoE workflow with validation}.}
\label{fig:full-moe-workflow}
\end{figure*}

\begin{figure*}[!h]
\centering
\begin{minipage}[t]{0.485\textwidth}
\textbf{Imperative Python/PyTorch baseline}
\begin{lstlisting}[style=pytorchcode]
from pathlib import Path
import json
import torch
from safetensors.torch import load_file, save_file

input_path = Path("(*@\bsio{models/input.safetensors}@*)")
source = load_file(str(input_path)) if input_path.suffix == ".safetensors" else torch.load(input_path, weights_only=True)
out = dict(source)

for layer in range((*@\bslit{16}@*)):
    prefix = f"model.layers.{layer}.mlp.experts"
    for proj in ("gate_proj", "up_proj", "down_proj"):
        e0 = f"{prefix}.0.{proj}.weight"
        e1 = f"{prefix}.1.{proj}.weight"
        delta = source[e1] - source[e0]
        u, s, vh = torch.linalg.svd(delta, full_matrices=False)
        sqrt_s = s[:(*@\bslit{64}@*)].sqrt()
        a = sqrt_s[:, None] * vh[:(*@\bslit{64}@*), :]
        b = u[:, :(*@\bslit{64}@*)] * sqrt_s
        out[f"{prefix}.1.{proj}.phlora_a.weight"] = a.(*@\pykey{to}@*)(
            (*@\pykey{dtype}@*)=torch.(*@\bslit{float16}@*), device=source[e1].device
        )
        out[f"{prefix}.1.{proj}.phlora_b.weight"] = b.(*@\pykey{to}@*)(
            (*@\pykey{dtype}@*)=torch.(*@\bslit{float16}@*), device=source[e1].device
        )
        (*@\pyop{del}@*) out[e1]

(*@\pyop{assert}@*) out["model.layers.0.mlp.experts.1.gate_proj.phlora_a.weight"](*@\pykey{.dtype}@*) == torch.(*@\bslit{float16}@*)
(*@\pyop{assert}@*) "model.layers.0.mlp.experts.1.gate_proj.weight" not in out

out_dir = Path("(*@\bsio{models/output}@*)")
max_bytes = (*@\bslit{1}@*) << (*@\bslit{30}@*)
out_dir.mkdir(parents=True, exist_ok=True)
shards, cur, cur_size = [], {}, (*@\bslit{0}@*)
for name, tensor in out.items():
    size = tensor.numel() * tensor.element_size()
    if cur and cur_size + size > max_bytes:
        shards.append(cur)
        cur, cur_size = {}, (*@\bslit{0}@*)
    cur[name] = tensor
    cur_size += size
if cur:
    shards.append(cur)

weight_map = {}
for idx, shard in enumerate(shards, start=(*@\bslit{1}@*)):
    shard_name = f"model-{idx:05d}-of-{len(shards):05d}.safetensors"
    save_file(shard, str(out_dir / shard_name))
    for name in shard:
        weight_map[name] = shard_name

(out_dir / "model.safetensors.index.json").write_text(
    json.dumps({"weight_map": weight_map}), encoding="utf-8"
)
\end{lstlisting}
\end{minipage}
\hfill
\begin{minipage}[t]{0.485\textwidth}
\textbf{\method{} plan}
\begin{lstlisting}[style=brainsurgeryyaml]
inputs:
- model::(*@\bsio{models/input.safetensors}@*)

transforms:
- copy:
    from: "(.*experts\\.1\\..*)\.weight"
    to: "\1.delta.weight"
- subtract_:
    from: "(.*experts)\\.0\\.(.*)\.weight"
    to: "\1.1.\2.delta.weight"
- phlora:
    target: "(.*experts\\.1\\..*)\.delta\.weight"
    target_a: "\1.phlora_a.weight"
    target_b: "\1.phlora_b.weight"
    rank: (*@\bslit{64}@*)
- cast_:
    target: ".*experts\\.1\\..*\.phlora_(a|b)\.weight"
    to: (*@\bslit{float16}@*)
- delete:
    target: ".*experts\.1\.(gate_proj|up_proj|down_proj)\.weight"
- assert:
    dtype:
      of: "model.layers.0.mlp.experts.1.gate_proj.phlora_a.weight"
      is: (*@\bslit{float16}@*)
- assert:
    not:
      exists: "model.layers.0.mlp.experts.1.gate_proj.weight"

output:
  path: (*@\bsio{models/output}@*)
  format: safetensors
  shard: (*@\bslit{1GB}@*)
\end{lstlisting}
\end{minipage}
\caption{\textbf{Full PHLoRA workflow with local validation}. The imperative baseline spells out checkpoint loading, PHLoRA factorization, casting, deletion, validation, and sharded export, while \method{} expresses the same workflow as one declarative plan.}
\label{fig:full-phlora-workflow}
\end{figure*}

\begin{figure*}[!h]
\centering
\begin{minipage}[t]{0.485\textwidth}
\textbf{Imperative Python/PyTorch baseline}
\begin{lstlisting}[style=pytorchcode]
from pathlib import Path
import json
import torch
from safetensors.torch import load_file, save_file

def load_checkpoint(path):
    return load_file(str(path)) if path.suffix == ".safetensors" else torch.load(path, weights_only=True)

def save_sharded_safetensors(sd, out_dir, max_bytes):
    out_dir.mkdir(parents=True, exist_ok=True)
    shards, cur, cur_size = [], {}, (*@\bslit{0}@*)
    for name, tensor in sd.items():
        size = tensor.numel() * tensor.element_size()
        if cur and cur_size + size > max_bytes:
            shards.append(cur)
            cur, cur_size = {}, (*@\bslit{0}@*)
        cur[name] = tensor
        cur_size += size
    if cur:
        shards.append(cur)
    weight_map = {}
    for idx, shard in enumerate(shards, start=(*@\bslit{1}@*)):
        shard_name = f"model-{idx:05d}-of-{len(shards):05d}.safetensors"
        save_file(shard, str(out_dir / shard_name))
        for name in shard:
            weight_map[name] = shard_name
    (out_dir / "model.safetensors.index.json").write_text(
        json.dumps({"weight_map": weight_map}), encoding="utf-8"
    )

def assert_same_state_dict(left, right):
    missing_l = sorted(set(right) - set(left))
    missing_r = sorted(set(left) - set(right))
    differing = [k for k in set(left) & set(right)
                 if left[k](*@\pykey{.shape}@*) != right[k](*@\pykey{.shape}@*)
                 or not (*@\pykey{torch.equal}@*)(left[k], right[k])]
    (*@\pyop{assert}@*) missing_l == missing_r == differing == []

source = load_checkpoint(Path("(*@\bsio{models/input.safetensors}@*)"))
ref = load_checkpoint(Path("(*@\bsio{models/low\_rank\_reference.safetensors}@*)"))
out = dict(source)

for layer in range((*@\bslit{16}@*)):
    for proj in ("gate_proj", "up_proj", "down_proj"):
        e0 = f"model.layers.{layer}.mlp.experts.0.{proj}.weight"
        e1 = f"model.layers.{layer}.mlp.experts.1.{proj}.weight"
        delta = source[e1] - source[e0]
        u, s, vh = torch.linalg.svd(delta, full_matrices=False)
        approx = (u[:, :(*@\bslit{64}@*)] * s[:(*@\bslit{64}@*)]) @ vh[:(*@\bslit{64}@*), :]
        out[e1] = (source[e0] + approx).(*@\pykey{to}@*)(
            (*@\pykey{dtype}@*)=torch.(*@\bslit{float16}@*),
            device=source[e1].device,
        )

(*@\pyop{assert}@*) out["model.layers.0.mlp.experts.1.gate_proj.weight"](*@\pykey{.dtype}@*) == torch.(*@\bslit{float16}@*)
assert_same_state_dict(out, ref)
save_sharded_safetensors(out, Path("(*@\bsio{models/low\_rank\_output}@*)"), (*@\bslit{1}@*) << (*@\bslit{30}@*))
\end{lstlisting}
\end{minipage}
\hfill
\begin{minipage}[t]{0.485\textwidth}
\textbf{\method{} plan}
\begin{lstlisting}[style=brainsurgeryyaml]
inputs:
  - model::(*@\bsio{models/input.safetensors}@*)
  - ref::(*@\bsio{models/low\_rank\_reference.safetensors}@*)

transforms:
  - subtract_:
      from: "model.layers\.(\d+)\.mlp\.experts\.0\.(.*_proj)\.weight"
      to: "model.layers.\1.mlp.experts.1.\2.weight"
  - phlora_:
      target: "model.layers\.(\d+)\.mlp\.experts\.1\.(.*_proj)\.weight"
      rank: (*@\bslit{64}@*)
  - add_:
      from: "model.layers\.(\d+)\.mlp\.experts\.0\.(.*_proj)\.weight"
      to: "model.layers.\1.mlp.experts.1.\2.weight"
  - cast_:
      target: "model.layers\.(\d+)\.mlp\.experts\.1\.(.*_proj)\.weight"
      to: (*@\bslit{float16}@*)
  - assert:
      dtype:
        of: "model.layers.0.mlp.experts.1.gate_proj.weight"
        is: (*@\bslit{float16}@*)
  - assert:
      all:
        - equal: { left: "model::(.+)", right: "ref::\1" }
        - equal: { left: "ref::(.+)", right: "model::\1" }

output:
  path: (*@\bsio{models/low\_rank\_output}@*)
  format: safetensors
  shard: (*@\bslit{1GB}@*)
\end{lstlisting}
\end{minipage}
\caption{\textbf{Full in-place low-rank expert rewrite with validation}.}
\label{fig:full-low-rank-workflow}
\end{figure*}

\section{Correctness Oracle and Preservation Scope}
\label{app:correctness-preservation}

\paragraph{Protocol.} We used two complementary correctness protocols. The controlled protocol applied ten transformations to deterministic safetensors fixtures constructed from literal and arithmetic-sequence tensors: identity serialization, copy, move, deletion, rename followed by inverse rename, split followed by concatenation, arithmetic operations, same-dtype conversion, FP32-to-BF16 conversion, and sharded save/reload. The real-checkpoint protocol applied an explicit multiply-by-one identity operation to one tensor in each of three pinned checkpoints: GPT-2 124M in a single safetensors file, OLMo 1B in indexed safetensors shards, and Pythia 1B in a single safetensors file.

\paragraph{Independent oracle.} \method{} was invoked only through its public command-line interface. A separate PyTorch/safetensors verifier that did not import \method{} constructed the expected state and compared the output tensor names, shapes, dtypes, and bytes. It also verified that every tensor outside the operation's declared write-set and every source checkpoint file remained byte-identical. The intentionally lossy FP32-to-BF16 case was required to match PyTorch's conversion exactly. Three negative controls changed a tensor value, a dtype, and the tensor-name set; the evaluation was valid only if the verifier detected all three corruptions. Table~\ref{tab:correctness-preservation} aggregates the two protocols.

\begin{table*}[t]
  \centering
  \small
  \begin{tabular}{lrrr}
    \toprule
    \textbf{Evaluation} & \textbf{Cases} & \textbf{Oracle exact $\uparrow$} & \textbf{Untouched exact $\uparrow$} \\
    \midrule
    Controlled primitives & 10/10 & 161/161 & 149/149 \\
    Pinned real checkpoints & 3/3 & 518/518 & 515/515 \\
    \midrule
    \textbf{Total} & \textbf{13/13} & \textbf{679/679} & \textbf{664/664} \\
    \bottomrule
  \end{tabular}
  \caption{Independent correctness and preservation checks ($\uparrow$ is better). Counts in the last two columns are tensor--case comparisons. Exact means identical names, shapes, dtypes, and tensor bytes. The real-checkpoint row covers GPT-2 124M, sharded OLMo 1B, and Pythia 1B at pinned revisions.}
  \label{tab:correctness-preservation}
\end{table*}

\paragraph{Results and scope.} Table~\ref{tab:correctness-preservation} shows that all 679/679 tensor--case comparisons matched the independently computed expected state and all 664/664 tensors outside the declared write-sets remained byte-identical. All source checkpoint files remained unchanged, and the verifier detected all three deliberately corrupted controls. We therefore observed neither an incorrect output tensor nor an unintended tensor change in these 13 cases. The evidence establishes exact tensor-state correctness only for the enumerated transformations, checkpoint layouts, and revisions. The oracle was independently implemented but author-written, not supplied by a third party. \method{} operates on tensor state dictionaries; custom safetensors header metadata and arbitrary checkpoint sidecars were outside the tested boundary and were not preserved. We therefore make no claim of complete container-level information preservation.

\section{Robustness and Failure Semantics}
\label{app:robustness}

\paragraph{Protocol.} We executed 18 deterministic cases through the public \method{} command-line interface. The cases exercised malformed plans, invalid transformations, zero or multiple tensor matches, invalid references, missing and damaged inputs, failed assertions, pre-existing destinations, injected save failure, and a process interruption. An auditor independent of \method{} hashed every source and destination file, parsed shard indexes, and loaded safetensors directly. For the three publication-fault cases, an evaluation-only wrapper raised an exception or paused the process immediately after one complete shard write; the interruption case was then terminated. The wrapper did not modify \method{} or affect the ordinary command-line cases.

\paragraph{Outcome criteria.} A case passed the evaluation only when its process outcome, diagnostic, source hashes, and observed destination state matched the frozen expectation. This evaluation-level pass is distinct from safe application behavior. A destination was classified as safe only when a failed operation published nothing or preserved a pre-existing destination byte-for-byte. Because publication behavior can depend on filesystem and process semantics, the cases were run on Linux and replicated on macOS; Table~\ref{tab:robustness} reports the Linux counts, without pooling the replication.

\begin{table*}[t]
  \centering
  \small
  \begin{tabular}{lrrr}
    \toprule
    \textbf{Scenario} & \textbf{Expected outcome $\uparrow$} & \textbf{Source unchanged $\uparrow$} & \textbf{Safe destination $\uparrow$} \\
    \midrule
    Before-save failures & 15/15 & 15/15 & 15/15 \\
    Fresh-dest. mid-save faults & 2/2 & 2/2 & 0/2 \\
    Existing-dest. mid-save fault & 1/1 & 1/1 & 0/1 \\
    Successful save control & 1/1 & 1/1 & 1/1 \\
    \midrule
    \textbf{Total} & \textbf{19/19} & \textbf{19/19} & \textbf{16/19} \\
    \bottomrule
  \end{tabular}
  \caption{Linux robustness and failure-semantics results, replicated without count pooling on macOS ($\uparrow$ is better). ``Expected outcome'' is the required success, failure, or interruption. A destination is safe when a failed operation publishes nothing or preserves an existing destination, or when the successful control publishes a complete loadable checkpoint.}
  \label{tab:robustness}
\end{table*}

\paragraph{Results and guidance.} Table~\ref{tab:robustness} shows that all 18 expected non-success outcomes occurred, all 17 applicable diagnostics matched, and all 18 source inputs remained byte-identical. Fourteen failures before publication exposed no destination, and one preserved a pre-existing destination exactly. The three mid-save cases revealed a limitation: save failure and process interruption left a visible shard without an index, while failure when targeting an existing checkpoint retained the old files and added an unindexed shard. Consequently, 16/18 cases met the destination-safety criterion. \method{} therefore does not currently provide atomic checkpoint publication. Outputs should be written to a fresh destination and accepted only after successful process exit and independent index/loading validation.

\section{Scaling Protocol and Complete Results}
\label{app:scaling}

\paragraph{Workload and models.} We evaluated one checkpoint-rewrite workload on ten pinned checkpoints spanning 70M--12B parameters, four model families, FP16/FP32/BF16 storage, and single-file and indexed-sharded inputs. The operation multiplied every floating-point tensor whose complete name matched \texttt{.*\textbackslash.weight} by 0.5, preserved every nonmatching tensor byte-for-byte, retained all names, shapes, and dtypes, and wrote safetensors with a 512~MiB shard budget. The four Pythia checkpoints form the within-family size series; the GPT-2, OLMo, and Qwen2.5 pairs test additional architectures and storage layouts.

\paragraph{Methods and measurement.} We compared an independent Python/PyTorch/safetensors implementation with \method{}'s in-memory and arena providers. All methods used CPU tensors, one I/O worker, identical inputs and storage, and one unmeasured warm-up followed by five warm-cache measured repetitions in rotating order on the same otherwise idle Linux host. Wall time covers process startup, checkpoint loading, transformation, and completed output saving. Peak RSS is the maximum sampled sum over the process tree. Effective throughput is the sum of logical input and output bytes divided by wall time. A measurement entered Table~\ref{tab:checkpoint-scaling} only after an independent oracle verified the complete output tensor set, values, dtypes, shard index, shard-size rule, and unchanged source files.

\begin{table*}[t]
\centering
\small
\resizebox{\textwidth}{!}{ 
\begin{tabular}{lllrrr}
\toprule
\textbf{Model} & \textbf{Method} & \textbf{Correct $\uparrow$} & \textbf{Wall (s) $\downarrow$} & \textbf{Peak RSS (GiB) $\downarrow$} & \textbf{Effective GiB/s $\uparrow$} \\
\midrule
Pythia 70M & Python/PyTorch & 5/5 & \textbf{1.17} & \textbf{0.77} & \textbf{0.26} \\
Pythia 70M & \method{} (memory) & 5/5 & 4.34 & 0.82 & 0.07 \\
Pythia 70M & \method{} (arena) & 5/5 & 4.38 & 0.88 & 0.07 \\
Pythia 410M & Python/PyTorch & 5/5 & \textbf{1.62} & 2.10 & \textbf{1.05} \\
Pythia 410M & \method{} (memory) & 5/5 & 5.23 & \textbf{1.52} & 0.32 \\
Pythia 410M & \method{} (arena) & 5/5 & 5.09 & 2.32 & 0.33 \\
Pythia 2.8B & Python/PyTorch & 5/5 & \textbf{4.30} & 10.96 & \textbf{2.46} \\
Pythia 2.8B & \method{} (memory) & 5/5 & 9.06 & \textbf{5.97} & 1.17 \\
Pythia 2.8B & \method{} (arena) & 5/5 & 9.26 & 11.21 & 1.14 \\
Pythia 12B & Python/PyTorch & 5/5 & \textbf{15.50} & 44.77 & \textbf{2.86} \\
Pythia 12B & \method{} (memory) & 5/5 & 24.59 & \textbf{22.88} & 1.81 \\
Pythia 12B & \method{} (arena) & 5/5 & 26.54 & 28.18 & 1.67 \\
GPT-2 124M & Python/PyTorch & 5/5 & \textbf{1.42} & 1.47 & \textbf{0.72} \\
GPT-2 124M & \method{} (memory) & 5/5 & 5.07 & \textbf{1.18} & 0.20 \\
GPT-2 124M & \method{} (arena) & 5/5 & 5.05 & 1.61 & 0.20 \\
GPT-2 XL 1.5B & Python/PyTorch & 5/5 & \textbf{5.02} & 12.29 & \textbf{2.38} \\
GPT-2 XL 1.5B & \method{} (memory) & 5/5 & 9.94 & \textbf{6.66} & 1.21 \\
GPT-2 XL 1.5B & \method{} (arena) & 5/5 & 10.77 & 12.50 & 1.11 \\
OLMo 1B (sharded) & Python/PyTorch & 5/5 & \textbf{4.51} & 9.65 & \textbf{2.11} \\
OLMo 1B (sharded) & \method{} (memory) & 5/5 & 9.70 & \textbf{5.44} & 0.98 \\
OLMo 1B (sharded) & \method{} (arena) & 5/5 & 9.35 & 9.37 & 1.02 \\
OLMo 7B (sharded, FP32) & Python/PyTorch & 5/5 & \textbf{18.40} & 30.81 & \textbf{2.79} \\
OLMo 7B (sharded, FP32) & \method{} (memory) & 5/5 & 24.98 & \textbf{26.33} & 2.05 \\
OLMo 7B (sharded, FP32) & \method{} (arena) & 5/5 & 30.23 & 29.02 & 1.70 \\
Qwen2.5 0.5B & Python/PyTorch & 5/5 & \textbf{1.79} & 2.34 & \textbf{1.03} \\
Qwen2.5 0.5B & \method{} (memory) & 5/5 & 5.39 & \textbf{1.59} & 0.34 \\
Qwen2.5 0.5B & \method{} (arena) & 5/5 & 5.36 & 2.35 & 0.34 \\
Qwen2.5 7B (sharded, BF16) & Python/PyTorch & 5/5 & \textbf{10.79} & 28.87 & \textbf{2.63} \\
Qwen2.5 7B (sharded, BF16) & \method{} (memory) & 5/5 & 16.09 & \textbf{14.86} & 1.76 \\
Qwen2.5 7B (sharded, BF16) & \method{} (arena) & 5/5 & 17.96 & 18.04 & 1.58 \\
\bottomrule
\end{tabular}
}
\caption{Controlled Linux CPU checkpoint rewrite ($\uparrow$/$\downarrow$ indicate better; bold marks the best time, memory, and throughput within each model). Values are medians over five correctness-validated repetitions; all 150/150 measured outputs passed the independent oracle.}
\label{tab:checkpoint-scaling}
\end{table*}

\paragraph{Results and scope.} All 150/150 measured outputs passed the oracle. Table~\ref{tab:checkpoint-scaling} shows that direct Python/PyTorch was faster at every point on the Pythia size series. At 12B, it required 15.50~s and 44.77~GiB peak RSS, compared with 24.59~s and 22.88~GiB for \method{}'s in-memory provider and 26.54~s and 28.18~GiB for the arena provider. The in-memory provider reduced peak RSS by 48.9\% relative to Python/PyTorch at 12B while taking 1.59$\times$ as long. These measurements demonstrate a time--memory trade-off, not a universal speed advantage. They apply to this single-process CPU checkpoint rewrite and do not establish GPU performance, distributed execution, optimizer-state support, multi-rank execution, or resharding.

\section{Additional System Comparison Experiments}
\label{app:competing-tools}

\paragraph{Matched operations.} We selected three operations with directly matching semantics on deterministic, real-shape-derived GPT-2 checkpoint fixtures. R01 performs regex/capture-based key rewriting and compares \method{} with \texttt{torch-state-bridge} 0.1.0. M01 performs a two-checkpoint weighted linear merge, and M02 performs base-relative two-task-vector arithmetic; both compare \method{} with MergeKit 0.1.4. Each pair received the same input tensors and the same independently specified output contract. R01 required byte-exact output; the two FP32 arithmetic cases used absolute and relative tolerances of $10^{-6}$.

\paragraph{Measurement and oracle.} The tools ran in isolated software environments on the same Linux host, using CPU execution, one thread, the same storage, one unmeasured warm-up, and five interleaved measured repetitions per tool--case pair. End-to-end wall time covers process startup, checkpoint loading, transformation, and completed output saving; peak RSS covers the process tree. An independent PyTorch/safetensors oracle, which imported none of the evaluated tools, checked tensor names, shapes, dtypes, and values before a measurement was admitted to Table~\ref{tab:competing-tools}.

\begin{table*}[t]
\centering
\small
\begin{tabular}{llrrrr}
\toprule
\textbf{Case} & \textbf{Tool} & \textbf{Correct $\uparrow$} & \textbf{Wall (s) $\downarrow$} & \textbf{Peak RSS (MiB) $\downarrow$} & \textbf{Output (MiB)} \\
\midrule
R01 & \method{} & \textbf{5/5} & 5.042 & 1785.7 & 522.7 \\
R01 & \texttt{torch-state-bridge} & \textbf{5/5} & \textbf{1.478} & \textbf{1571.9} & 522.7 \\
M01 & \method{} & \textbf{5/5} & 5.292 & \textbf{2166.7} & 474.7 \\
M01 & MergeKit & \textbf{5/5} & \textbf{4.887} & 2890.4 & 474.7 \\
M02 & \method{} & \textbf{5/5} & 5.480 & \textbf{3115.1} & 474.7 \\
M02 & MergeKit & \textbf{5/5} & \textbf{5.289} & 3895.3 & 474.7 \\
\bottomrule
\end{tabular}
\caption{Controlled checkpoint-operation comparison on Linux ($\uparrow$/$\downarrow$ indicate better; bold marks the best value within each case, with correctness ties). Runtime and peak RSS are medians over five correctness-validated repetitions per tool and case. End-to-end time includes process startup, checkpoint loading, transformation, and output saving.}
\label{tab:competing-tools}
\end{table*}

\paragraph{Results and scope.} All 30/30 measured outputs passed the common oracle. For R01, \method{} required 5.042~s and 1785.7~MiB peak RSS, versus 1.478~s and 1571.9~MiB for \texttt{torch-state-bridge}: \method{} was 3.41$\times$ slower and used 13.6\% more peak memory. For M01 and M02, the \method{}-to-MergeKit median wall-time ratios were 1.083 and 1.036, while \method{} used 25.0\% and 20.0\% less peak RSS. The MergeKit cases therefore showed similar runtime and lower peak memory for \method{}, whereas R01 showed a clear disadvantage to \texttt{torch-state-bridge}. These three matched operations do not establish an overall tool ranking or a usability advantage. MergeKit slicing, PEFT, Orbax, and PyTorch Distributed Checkpoint provide additional functionality that was not evaluated by these like-for-like cases.

\section{Behavioral Preservation}
\label{app:behavioral-analysis}

\paragraph{Prompt suite.} We evaluate behavioral preservation on a frozen, deterministically selected suite of 70 prompts: 30 multilingual reading-comprehension prompts from Belebele~\citep{bandarkar-etal-2024-belebele}, 30 knowledge and reasoning prompts from MMLU~\citep{hendrycks-etal-2021-mmlu}, and 10 Python code-completion prompts from HumanEval~\citep{chen-etal-2021-evaluating}. The manifest records each source revision, upstream identifier, row, language or subject stratum, license, rendered prompt, and checksum. Selection used a frozen SHA-256 ranking and was independent of model outputs. This is a behavioral-regression diagnostic, not a claim about general task accuracy, multilingual competence, HumanEval pass rate, or safety.

\paragraph{Protocol and metrics.} Table~\ref{tab:behavioral-methodology} contains two distinct evaluations. The structural-losslessness panel applies a value-preserving namespace move followed by exact name restoration, without arithmetic or dtype conversion, to ten checkpoints at their native evaluated dtypes. The Pythia precision-ablation panel instead multiplies each selected weight by 0.5 and then by 2, comparing native FP16 with FP32 storage and inference throughout. For both panels, the reference is the checkpoint before the round trip and the transformed model is the checkpoint after it. ``Exact tensors'' counts byte-identical transformed/reference tensor pairs; PPL ratio is transformed/reference mean perplexity; final and sequence cosine are mean logit-vector cosine similarities at the final prompt position and across all compared sequence positions; maximum difference is the largest absolute full-sequence logit difference; exact generation requires all 32 greedy token IDs to match; and MCQ agreement compares predictions on the 60 Belebele and MMLU prompts. Each model--dtype condition used the same 70 prompts, greedy decoding, deterministic inference settings, the checkpoint's pinned tokenizer, and one NVIDIA B200.

\begin{table*}[t]
\centering
\scriptsize
\resizebox{\textwidth}{!}{
\begin{tabular}{llrrrrrrrrrr}
\toprule
\multicolumn{12}{l}{\textbf{Structural losslessness}} \\
\textbf{Model} & \textbf{Dtype} & \textbf{N} & \textbf{Exact tensors $\uparrow$} & \textbf{PPL ratio $\rightarrow 1$} & \textbf{Final cos. $\uparrow$} & \textbf{Seq. cos. $\uparrow$} & \textbf{Max diff $\downarrow$} & \textbf{Exact gen. $\uparrow$} & \textbf{MCQ agree. $\uparrow$} & \textbf{NaN $\downarrow$} & \textbf{Inf $\downarrow$} \\
\midrule
Pythia 70M & float16 & 70 & 94/94 & 1.00000000 & 0.99998998 & 1.00000000 & 0 & 70/70 & 60/60 & 0 & 0 \\
Pythia 410M & float16 & 70 & 364/364 & 1.00000000 & 0.99999699 & 1.00000000 & 0 & 70/70 & 60/60 & 0 & 0 \\
Pythia 2.8B & float16 & 70 & 484/484 & 1.00000000 & 0.99999698 & 1.00000000 & 0 & 70/70 & 60/60 & 0 & 0 \\
Pythia 12B & float16 & 70 & 544/544 & 1.00000000 & 0.99999690 & 1.00000000 & 0 & 70/70 & 60/60 & 0 & 0 \\
GPT-2 124M & float32 & 70 & 160/160 & 1.00000000 & 0.99999984 & 1.00000000 & 0 & 70/70 & 60/60 & 0 & 0 \\
GPT-2 XL 1.5B & float32 & 70 & 628/628 & 1.00000000 & 0.99999986 & 1.00000000 & 0 & 70/70 & 60/60 & 0 & 0 \\
OLMo 1B (sharded) & float32 & 70 & 114/114 & 1.00000000 & 0.99999992 & 1.00000000 & 0 & 70/70 & 60/60 & 0 & 0 \\
OLMo 7B (sharded, FP32) & float32 & 70 & 226/226 & 1.00000000 & 0.99999990 & 1.00000000 & 0 & 70/70 & 60/60 & 0 & 0 \\
Qwen2.5 0.5B & bfloat16 & 70 & 290/290 & 1.00000000 & 0.99999174 & 1.00000000 & 0 & 70/70 & 60/60 & 0 & 0 \\
Qwen2.5 7B (sharded, BF16) & bfloat16 & 70 & 339/339 & 1.00000000 & 0.99999312 & 1.00000000 & 0 & 70/70 & 60/60 & 0 & 0 \\
\midrule
\multicolumn{12}{l}{\textbf{Pythia precision ablation}} \\
\textbf{Model} & \textbf{Dtype} & \textbf{N} & \textbf{Exact tensors $\uparrow$} & \textbf{PPL ratio $\rightarrow 1$} & \textbf{Final cos. $\uparrow$} & \textbf{Seq. cos. $\uparrow$} & \textbf{Max diff $\downarrow$} & \textbf{Exact gen. $\uparrow$} & \textbf{MCQ agree. $\uparrow$} & \textbf{NaN $\downarrow$} & \textbf{Inf $\downarrow$} \\
\midrule
Pythia 70M (native FP16) & float16 & 70 & 68/94 & 1.01559670 & 0.99998982 & 0.99999995 & 12.5 & 34/70 & 60/60 & 0 & 0 \\
Pythia 70M (FP32 throughout) & float32 & 70 & 94/94 & 1.00000000 & 0.99999982 & 1.00000000 & 0 & 70/70 & 60/60 & 0 & 0 \\
Pythia 410M (native FP16) & float16 & 70 & 266/364 & 1.00173069 & 0.99963612 & 0.99980833 & 2.3691406 & 59/70 & 58/60 & 0 & 0 \\
Pythia 410M (FP32 throughout) & float32 & 70 & 364/364 & 1.00000000 & 0.99999991 & 1.00000000 & 0 & 70/70 & 60/60 & 0 & 0 \\
Pythia 2.8B (native FP16) & float16 & 70 & 354/484 & 1.00102777 & 0.99989628 & 0.99989772 & 1.8027344 & 64/70 & 58/60 & 0 & 0 \\
Pythia 2.8B (FP32 throughout) & float32 & 70 & 484/484 & 1.00000000 & 0.99999995 & 1.00000000 & 0 & 70/70 & 60/60 & 0 & 0 \\
Pythia 12B (native FP16) & float16 & 70 & 398/544 & 1.00223011 & 0.99962275 & 0.99959205 & 2.390625 & 49/70 & 57/60 & 0 & 0 \\
Pythia 12B (FP32 throughout) & float32 & 70 & 544/544 & 1.00000000 & 0.99999992 & 1.00000000 & 0 & 70/70 & 60/60 & 0 & 0 \\
\bottomrule
\end{tabular}
}
\caption{Behavioral preservation under a value-preserving structural round trip and a separate Pythia arithmetic precision ablation. $N$ is the number of prompts per checkpoint. PPL ratio is transformed/reference and is optimal at 1. Exact generation requires all 32 generated token IDs to agree; MCQ agreement covers the 60 Belebele and MMLU prompts. Cosines are computed numerically, so values can round slightly below 1 even when the corresponding maximum absolute difference is zero.}
\label{tab:behavioral-methodology}
\end{table*}

\paragraph{Results and scope.} The structural round trip preserved every tensor exactly across all ten checkpoints and produced a PPL ratio of 1.0, zero maximum logit difference, 70/70 exact generations, 60/60 MCQ agreement, and no NaN or Inf values. The separate Pythia arithmetic ablation produced restoration differences under native FP16 but exact tensor and behavioral restoration under FP32. The ablation therefore shows that an algebraically inverse arithmetic sequence need not be numerically reversible in FP16; it does not contradict the losslessness of the non-arithmetic structural round trip. These 70 prompts are a behavioral-regression diagnostic and do not establish general task accuracy, multilingual competence, code correctness, or safety.

\section{Coding-Agent Implementation and Auditability Evaluation}
\label{app:coding-agent-evaluation}

\paragraph{Research questions.} We evaluate \method{} as an interface for implementing the five checkpoint transformations summarized in Table~\ref{tab:agent-tasks}, rather than as a source of pretrained-model quality improvements. The evaluation asks whether coding agents (i) produce functionally correct artifacts, (ii) succeed on their first execution and avoid execution failures, (iii) require less authored implementation code, and (iv) produce implementations that are more structurally uniform across model architectures. We additionally measure end-to-end solution time, token use, monetary cost, and the ability to identify defects in an independently supplied artifact. Coding agents serve as controlled proxy practitioners; this experiment is not a human-subject usability study.

\paragraph{Factorial design.} Each run is one fresh agent session for one checkpoint-editing task, target model, implementation condition, effort tier, and repeat. The design crosses five tasks, three target models, three implementation conditions, and three vendor-specific effort tiers. Runs have no conversational state from other cells, and a new sandbox is created for every cell. Repeats use new sessions and sandboxes; no artifact produced in one cell is available in another. Cell-level records retain every factor; the paper reports agent--effort--condition strata and paired comparisons that preserve task and target grouping.

\begin{table*}[t]
\centering
\small
\begin{tabular}{lll}
\toprule
\textbf{Task} & \textbf{Operation} & \textbf{Capability exercised} \\
\midrule
T1 & Depth pruning & bulk deletion, collision-safe renaming, structural assertions \\
T2 & Attention-head pruning & architecture-dependent slicing, concatenation, shape assertions \\
T3 & Mixed-precision export & dtype conversion, selective deletion, sharded safetensors output \\
T4 & Task-vector merge & multi-checkpoint arithmetic and equality preconditions \\
T5 & LoRA merge & low-rank matrix arithmetic, transposition, and sharded output \\
\bottomrule
\end{tabular}
\caption{Checkpoint-editing tasks used in the coding-agent evaluation. Every task specifies an output contract and required executable checks independently of the implementation condition.}
\label{tab:agent-tasks}
\end{table*}

\paragraph{Target models.} Every task is instantiated for GPT-2 124M, OLMo-1B-0724-hf, and Pythia-1B at pinned revisions. These targets differ in checkpoint layout and tensor organization: GPT-2 uses fused QKV projections and Conv1D matrix layout in a single FP32 file; OLMo uses separate attention projections and an indexed-sharded FP32 checkpoint; and Pythia uses interleaved fused QKV projections in FP16. The model-specific task specifications enumerate the required names, shapes, dtypes, counts, and output layout. All conditions receive the same target-specific specification and read-only inputs.

\paragraph{Implementation conditions.} In the Python condition (P), the agent writes a standalone script using pinned PyTorch and safetensors packages and may not use \method{}. In the free-choice condition (F), the agent may use a frozen allow-list comprising PyTorch, safetensors, Transformers, PEFT, MergeKit, torch-state-bridge, Accelerate, and Hugging Face Hub, but not \method{}; the selected tools and versions are recorded. In the \method{} condition (B), the agent must express the complete transformation and its required checks as a \method{} YAML plan and may not write Python. It receives a frozen documentation pack containing the public README, interface reference, complete command help, and two unrelated example plans. \method{} is installed without exposing its repository source inside the participant sandbox.

\paragraph{Agents and inference settings.} The Claude Code cohort uses Claude Sonnet~5, Claude Opus~5, and Claude Fable~5.1 at the vendor's low, medium, and high effort settings. The Codex cohort uses GPT-5.6~Sol at light, medium, and high study tiers, corresponding to recorded low, medium, and high reasoning-effort settings. Each agent contributes 270 solve cells: three targets, five tasks, three conditions, three effort levels, and two repeats. Claude cells have a 40-turn and 1,800-second solve cap; Codex cells have the same 1,800-second time cap and no artificial turn cap. One Sonnet cell records the earlier 900-second setting; it did not hit the cap. No evaluated cell hits a time or turn cap. Exact provider model identifiers, repository revisions, environment freezes, and rate-card inputs are retained with the run records.

\paragraph{Isolation and reproducibility.} Every cell runs in a fresh working directory and private environment, and all checkpoint inputs are read-only. Claude's sandbox policy denies web access, package installation, and reads outside the cell; Codex runs in its workspace-write sandbox with network access disabled. The three conditions use pinned Torch~2.14.0, safetensors~0.5.3, and NumPy~2.5.2 environments; condition-specific dependencies and the \method{} revision are recorded with each run. A cross-machine manifest verifies the pinned base checkpoints, generated inputs, hidden references, and documentation pack before a cohort begins. Runs are excluded if this manifest does not verify in full. The task prompts, condition instructions, graders, references, and defect fixtures are frozen before evaluated runs begin.

\paragraph{Independent functional grading.} After the participant session terminates, a grader that imports neither \method{} nor the comparison tools evaluates the produced checkpoint against a hidden PyTorch/safetensors reference. It checks loadability, exact tensor-name sets, shapes, dtypes, values, and task-specific sharding requirements. Comparisons are byte-exact except for tensors created by floating-point arithmetic: FP32 arithmetic outputs use a relative Frobenius tolerance of $10^{-5}$ and half-precision arithmetic outputs use $10^{-3}$. Final success is the fraction of graded artifacts that satisfy every check; an absent or incomplete artifact is a failure.

\paragraph{Execution reliability.} The runner records every execution of the participant's script, selected tool, or \method{} plan, its process outcome, and the execution number at which the first success occurs. We report first-execution failure rate, the fraction of runs containing any failed execution, failed executions divided by all executions, and executions until first success. Additional successful executions are not automatically counted as errors or repairs because an agent may rerun a valid artifact for verification. Failed executions are manually assigned a frozen category, including crash or parse failure, failed assertion, no match, overmatch, existing destination, shape/dtype error, save error, partial output, or other. Shell and harness failures are reported separately from transformation-language or implementation failures.

\paragraph{Implementation size.} Authored implementation size is the number of non-blank, non-comment lines in the final executable or declarative artifact. For a condition-F solution containing multiple authored files, the count is summed across its scripts and configuration files. Reports, logs, generated checkpoints, package source, and unchanged supplied files are excluded. Shared helper code is reported separately and is not silently removed from only one condition. We present paired \method{}--Python differences for the same agent, effort tier, repeat, task, and target, together with task-level results; this prevents a single verbose task from being hidden by an overall average.

\paragraph{Implementation convergence.} Similarity is computed only among independently authored implementations of the same task, target model, and condition; artifacts for unrelated tasks or targets are never compared. We remove blank and comment-only lines and replace filesystem-path literals with a common placeholder before tokenization, while retaining tensor names, shapes, slice indices, dtypes, operators, assertions, and sharding choices. For each eligible pair, we compute the Jaccard similarity of normalized token 5-gram sets. Scores range from zero to one, where one denotes identical normalized 5-gram sets. This metric measures convergence of implementation form rather than functional correctness, readability, or maintainability; those properties are evaluated or discussed separately.

\paragraph{Time, tokens, and cost.} Solve time is measured from participant-process launch until termination and therefore includes reading the supplied task and documentation, authoring, executions, debugging, and the final report. This cold-start, end-to-end duration is the primary time measurement. For Claude, complete transcripts additionally permit a secondary decomposition that subtracts intervals attributed to consultation of the supplied \method{} documentation; this is reported separately and does not replace total solve time. The transcript parser does not provide the corresponding measure for Codex, so no cross-vendor documentation-adjusted comparison is made. Input, cache-read, cache-write, output, and reasoning tokens are retained separately. Provider-reported cost is used when available; otherwise cost is computed from a frozen rate card recorded with the run.

\paragraph{Artifact review.} After solving, a fresh single-turn session of the same agent receives the task specification and one independently supplied artifact and must decide whether it satisfies the specification and explain any rejection. Odd repeats contain an artifact with one frozen injected defect, whereas even repeats contain the correct reference artifact. The primary score follows the leading YES/NO verdict: NO on a defective artifact is a defect rejection, and NO on a correct artifact is a false alarm. Free-text explanations are retained but are not separately scored for exact defect identification. Defect-rejection and false-alarm rates are reported separately; defective-only evaluations are not treated as complete auditability evidence.

\paragraph{Analysis and reporting.} The run is the observational unit. Counts accompany every proportion, and condition contrasts are paired on agent, effort, repeat, task, and target. Tables~\ref{tab:agent-implementation}--\ref{tab:agent-auditability} report the principal stratified results; the artifact package retains cell-level outcomes, failure logs, tokens, normalization rules, manifests, and exclusions. We interpret the evaluation as evidence about implementation reliability, concision, form convergence, and artifact review for the tested tasks and agents, not universal speed, human usability, or downstream model quality.

\begin{table*}[t]
\centering
\small
\resizebox{\textwidth}{!}{%
\begin{tabular}{llccccc}
\toprule
\textbf{Agent} & \textbf{Effort} & \textbf{One-shot (\%) $\uparrow$} & \textbf{First failures $\downarrow$} & \textbf{Time (s) $\downarrow$} & \textbf{Cost (USD) $\downarrow$} & \textbf{LOC $\downarrow$} \\
\multicolumn{2}{c}{} & \multicolumn{5}{c}{\textit{B / P / F within each cell}} \\
\midrule
\multicolumn{7}{l}{\textbf{Claude Code}} \\
Claude Fable 5.1 & Low & 100.0 / 90.0 / 93.3 & 0 / 3 / 2 & 88.1 / 46.7 / 56.4 & 0.797 / 0.479 / 0.532 & 31.0 / 65.5 / 54.5 \\
Claude Fable 5.1 & Medium & 100.0 / 100.0 / 93.3 & 0 / 0 / 2 & 90.7 / 50.1 / 70.7 & 0.968 / 0.511 / 0.629 & 33.0 / 77.5 / 83.0 \\
Claude Fable 5.1 & High & 100.0 / 93.3 / 100.0 & 0 / 2 / 0 & 132.8 / 68.7 / 95.9 & 1.309 / 0.643 / 0.781 & 42.0 / 111.0 / 106.5 \\
Claude Opus 5 & Low & 100.0 / 100.0 / 96.7 & 0 / 0 / 1 & 92.1 / 49.5 / 67.6 & 0.594 / 0.301 / 0.359 & 34.0 / 69.0 / 68.5 \\
Claude Opus 5 & Medium & 100.0 / 96.7 / 93.3 & 0 / 1 / 2 & 114.1 / 69.7 / 109.7 & 0.744 / 0.372 / 0.538 & 45.5 / 94.5 / 99.0 \\
Claude Opus 5 & High & 100.0 / 100.0 / 86.7 & 0 / 0 / 4 & 210.8 / 125.5 / 175.1 & 1.342 / 0.573 / 0.813 & 56.5 / 133.5 / 136.5 \\
Claude Sonnet 5 & Low & 96.7 / 93.3 / 90.0 & 1 / 2 / 3 & 119.4 / 54.5 / 68.0 & 0.306 / 0.139 / 0.173 & 37.5 / 91.0 / 89.5 \\
Claude Sonnet 5 & Medium & 96.7 / 100.0 / 86.7 & 1 / 0 / 4 & 160.8 / 60.2 / 83.6 & 0.463 / 0.157 / 0.207 & 45.5 / 102.0 / 99.5 \\
Claude Sonnet 5 & High & 100.0 / 93.3 / 90.0 & 0 / 2 / 3 & 230.3 / 81.6 / 114.9 & 0.688 / 0.196 / 0.256 & 52.0 / 105.5 / 105.0 \\
\addlinespace
\multicolumn{7}{l}{\textbf{Codex}} \\
GPT-5.6 Sol & Low (light) & 60.0 / 100.0 / 96.7 & 12 / 0 / 1 & 150.9 / 97.7 / 131.7 & 0.434 / 0.230 / 0.281 & 39.5 / 73.0 / 82.5 \\
GPT-5.6 Sol & Medium & 73.3 / 100.0 / 93.3 & 8 / 0 / 2 & 176.2 / 131.9 / 169.3 & 0.483 / 0.287 / 0.342 & 49.0 / 93.0 / 95.5 \\
GPT-5.6 Sol & High & 83.3 / 96.7 / 76.7 & 5 / 1 / 7 & 220.5 / 149.1 / 212.6 & 0.561 / 0.317 / 0.377 & 52.5 / 92.5 / 132.5 \\
\midrule
\multicolumn{7}{l}{\textbf{Implementation convergence among solutions to the same task and target}} \\
\multicolumn{2}{l}{\textbf{Condition}} & \multicolumn{5}{l}{\textbf{Median normalized token 5-gram Jaccard [IQR]}} \\
\multicolumn{2}{l}{\method{} (B)} & \multicolumn{5}{l}{0.348 [0.248, 0.469]} \\
\multicolumn{2}{l}{Python/PyTorch (P)} & \multicolumn{5}{l}{0.075 [0.052, 0.117]} \\
\multicolumn{2}{l}{Free choice (F)} & \multicolumn{5}{l}{0.064 [0.043, 0.100]} \\
\bottomrule
\end{tabular}%
}
\caption{Complete implementation-phase results by agent and effort. Each B/P/F entry summarizes 30 sessions: three targets, five tasks, and two repeats. All final artifacts pass the independent grader. One-shot denotes success on the first execution; time, cost, and nonblank, non-comment LOC are medians. Similarity is computed only among normalized solutions to the same task and target.}
\label{tab:agent-implementation}
\end{table*}

\paragraph{Implementation results.}
All 1,080 final artifacts pass and no session reaches a cap, so eventual completion does not distinguish the conditions. First-execution reliability does: Claude favors B over P by $+3.0$ points, while Sol favors P by $26.7$ points (Table~\ref{tab:agent-paired}). Across Claude sessions, B/P/F contain failed attempts in 7/10/24 sessions (10/10/24 failed attempts); the corresponding Sol counts are 22/1/4 sessions (26/1/4 attempts). BrainSurgery plans are shorter and more structurally convergent, but fresh-session time, tokens, and cost are higher. Median Claude input/output tokens are 533,948/6,616 for B, 172,694/3,812 for P, and 242,996/5,262 for F; Sol medians are 386,938/3,691, 167,585/2,768, and 211,819/4,125. Subtracting transcript-derived documentation consultation reduces the Claude B/P time ratio from 2.10 to 1.43 (95\% CI $[1.27,1.58]$); Codex transcripts do not support the same decomposition.

\begin{table*}[t]
\centering
\scriptsize
\setlength{\tabcolsep}{4pt}
\begin{tabular}{llrrrr}
\toprule
\textbf{Scope} & \textbf{Contrast} & \textbf{One-shot diff. (pp)} & \textbf{Time ratio} & \textbf{Cost ratio} & \textbf{LOC ratio} \\
\midrule
Claude agents & B/P & $+3.0\ [ +1.1,+5.2]$ & $2.10\ [1.78,2.41]$ & $2.01\ [1.77,2.22]$ & $0.43\ [0.35,0.65]$ \\
Claude agents & B/F & $+7.0\ [ +3.3,+11.1]$ & $1.53\ [1.27,1.76]$ & $1.53\ [1.38,1.76]$ & $0.43\ [0.35,0.65]$ \\
GPT-5.6 Sol & B/P & $-26.7\ [-40.0,-14.4]$ & $1.39\ [1.27,1.74]$ & $1.78\ [1.60,2.01]$ & $0.55\ [0.39,0.72]$ \\
GPT-5.6 Sol & B/F & $-16.7\ [-31.1,-3.3]$ & $1.02\ [0.90,1.30]$ & $1.47\ [1.31,1.67]$ & $0.47\ [0.35,0.62]$ \\
All agents & B/P & $-4.4\ [-7.5,-1.4]$ & $1.95\ [1.69,2.32]$ & $2.13\ [1.94,2.32]$ & $0.46\ [0.36,0.65]$ \\
All agents & B/F & $+1.1\ [-2.8,+5.3]$ & $1.31\ [1.17,1.56]$ & $1.68\ [1.52,1.86]$ & $0.44\ [0.35,0.63]$ \\
\bottomrule
\end{tabular}
\caption{Paired implementation-condition contrasts with 95\% cluster-bootstrap intervals over the 15 target--task units. Ratios below one favor B for time, cost, and LOC. Agent families are reported separately because the first-execution interaction makes the pooled reliability contrast misleading.}
\label{tab:agent-paired}
\end{table*}

\begin{table*}[t]
\centering
\small
\resizebox{\textwidth}{!}{%
\begin{tabular}{llrrrrrr}
\toprule
\textbf{Agent} & \textbf{Effort} & \textbf{Defects rejected $\uparrow$} & \textbf{Correct accepted $\uparrow$} & \textbf{Correct rejected $\downarrow$} & \textbf{Bal. acc. (\%) $\uparrow$} & \textbf{Time (s) $\downarrow$} & \textbf{Cost (USD) $\downarrow$} \\
\midrule
\multicolumn{8}{l}{\textbf{Claude Code}} \\
Claude Fable 5.1 & Low & 45/45 & 45/45 & 0/45 & 100.0 & 6.2 & 0.218 \\
Claude Fable 5.1 & Medium & 45/45 & 45/45 & 0/45 & 100.0 & 8.1 & 0.224 \\
Claude Fable 5.1 & High & 45/45 & 45/45 & 0/45 & 100.0 & 8.8 & 0.230 \\
Claude Opus 5 & Low & 45/45 & 45/45 & 0/45 & 100.0 & 6.5 & 0.112 \\
Claude Opus 5 & Medium & 45/45 & 45/45 & 0/45 & 100.0 & 9.1 & 0.118 \\
Claude Opus 5 & High & 45/45 & 45/45 & 0/45 & 100.0 & 13.8 & 0.131 \\
Claude Sonnet 5 & Low & 45/45 & 32/45 & 13/45 & 85.6 & 4.9 & 0.046 \\
Claude Sonnet 5 & Medium & 45/45 & 37/45 & 8/45 & 91.1 & 7.8 & 0.049 \\
Claude Sonnet 5 & High & 45/45 & 42/45 & 3/45 & 96.7 & 13.1 & 0.056 \\
\addlinespace
\multicolumn{8}{l}{\textbf{Codex}} \\
GPT-5.6 Sol & Low (light) & 45/45 & 33/45 & 12/45 & 86.7 & 8.2 & 0.059 \\
GPT-5.6 Sol & Medium & 45/45 & 35/45 & 10/45 & 88.9 & 8.2 & 0.060 \\
GPT-5.6 Sol & High & 45/45 & 35/45 & 10/45 & 88.9 & 8.7 & 0.060 \\
\midrule
\multicolumn{8}{l}{\textbf{Condition summary across all agents and effort levels}} \\
\method{} (B) & All & 180/180 & 159/180 & 21/180 & 94.2 & 9.9 & 0.109 \\
Python/PyTorch (P) & All & 180/180 & 162/180 & 18/180 & 95.0 & 7.5 & 0.104 \\
Free choice (F) & All & 180/180 & 163/180 & 17/180 & 95.3 & 7.5 & 0.105 \\
\bottomrule
\end{tabular}%
}
\caption{\textbf{Artifact-review results by agent, vendor-defined effort, and implementation condition.} Each agent--effort row contains 45 defective and 45 correct artifacts; each condition-summary row contains 180 of each. Defect rejection counts NO verdicts on defective artifacts; correct acceptance and rejection count YES and NO verdicts on correct artifacts. Balanced accuracy averages defect rejection and correct acceptance. Codex calls its low tier light. Values follow the leading YES/NO verdict; free-text explanations are not separately scored for exact defect identification.}
\label{tab:agent-auditability}
\end{table*}

\paragraph{Auditability results.} Reviewers reject every injected defect (540/540) and accept 484/540 correct artifacts, yielding 94.8\% balanced accuracy overall. Condition-level values are 94.2\% for B, 95.0\% for P, and 95.3\% for F; the paired B--P difference is $-1.0$ points (95\% cluster-bootstrap CI $[-2.9,+0.7]$). P and F are separate review sessions over the same Python artifact representation, so they are not a three-tool comparison. All 56 false alarms come from Sonnet~5 and Sol and concentrate in task-vector merging (33/108 correct artifacts rejected) and LoRA merging (15/108), followed by mixed-precision export (5/108), layer pruning (3/108), and head pruning (0/108). The evidence therefore supports high absolute review accuracy for the evaluated plans but no representation-level advantage; descriptively, the larger differences occur across reviewers and tasks rather than among the three condition-level averages.

\subsection{Cold-start Time Test}
\label{app:cold-start-time-test}

We compare cold-start time, including reading the supplied materials, implementation, execution, debugging, and reporting, using the ratio between median \method{} and Python/PyTorch times within each agent cohort. The ratios are 2.10 for Claude and 1.39 for Sol, with corresponding cost ratios are 2.01 and 1.78, showing that \method{} currently requires more initial time and cost.

\section{Full Results on System Comparison}
\label{app:full-res-system-comparison}

\subsection{\method{} Plans Length}
For each system and case analyzed we implement a \method{} YAML plan. One system considered (LLM-Pruner) implement special algorithms for determining model edits to apply. In this last case, depending on the algorithm, a high number of single-neuron operations may be necessary, for this reason plans may reach up to around 160.000 rows. Importantly, coding-agents showed high performances in generating plans (\ref{sec:agent-evaluation}), and in this case agents were used to map algorithm decisions to plans in a fast and reliable way. In other cases plans are very small and vary from 15 (All MergeKit cases) to 65 (LLM Neuronatomy) rows.

In such   and  For cases in which the model edits are not determined by a custom algorithm which determines  

\begin{table*}[t]
\centering
\scriptsize
\setlength{\tabcolsep}{4pt}
\renewcommand{\arraystretch}{1.15}
\begin{tabular}{ll ccc}
\toprule
\textbf{Task} & \textbf{Metric} & \textbf{block\_l2} & \textbf{block\_taylor} & \textbf{layer\_wise} \\
\midrule
\multirow{2}{*}{ARC-Challenge}
& acc      & 25.77 & 31.91 & 33.70 \\
& acc\_norm & 28.50 & 36.52 & 35.75 \\
\addlinespace
\multirow{2}{*}{ARC-Easy}
& acc      & 32.87 & 61.03 & 53.79 \\
& acc\_norm & 32.58 & 50.04 & 54.00 \\
\addlinespace
BoolQ & acc & 61.68 & 56.85 & 74.19 \\
\addlinespace
\multirow{2}{*}{HellaSwag}
& acc      & 31.10 & 50.71 & 39.30 \\
& acc\_norm & 37.06 & 66.82 & 51.28 \\
\addlinespace
\multirow{2}{*}{OpenBookQA}
& acc      & 16.80 & 29.40 & 28.40 \\
& acc\_norm & 29.80 & 40.00 & 35.20 \\
\addlinespace
\multirow{2}{*}{PIQA}
& acc      & 57.94 & 75.68 & 66.00 \\
& acc\_norm & 57.29 & 75.08 & 66.32 \\
\addlinespace
WinoGrande & acc & 52.33 & 59.59 & 59.67 \\
\bottomrule
\end{tabular}
\caption{LLM-Pruner suite: full downstream evaluation results (\%). Values shown are identical across LLM-Pruner (reference), \method{} arena, and \method{} in-memory: a paired bootstrap comparison (10{,}000 resamples, seed 0) over matched per-example predictions gives delta $=[+0.0000, +0.0000]$ for every metric, every case, and both execution modes ($n$ ranging from 500 matched examples for OpenBookQA to 10{,}042 for HellaSwag). WinoGrande is evaluated for all three cases.}
\label{tab:downstream-pruner-full}
\end{table*}

\begin{table*}[t]
\centering
\scriptsize
\setlength{\tabcolsep}{3pt}
\renewcommand{\arraystretch}{1.1}
\resizebox{\textwidth}{!}{%
\begin{tabular}{ll ccc ccc ccc}
\toprule
& & \multicolumn{3}{c}{\textbf{MMLU (\%)}} & \multicolumn{3}{c}{\textbf{HellaSwag acc (\%)}} & \multicolumn{3}{c}{\textbf{HellaSwag acc\_norm (\%)}} \\
\cmidrule(lr){3-5} \cmidrule(lr){6-8} \cmidrule(lr){9-11}
\textbf{Model} & \textbf{Merge type} & MergeKit & Arena & In-memory & MergeKit & Arena & In-memory & MergeKit & Arena & In-memory \\
\midrule
\multirow{3}{*}{olmo2}
& linear          & $74.31_{[73.58,75.05]}$ & $73.49_{[72.76,74.21]}$ & $73.49_{[72.71,74.23]}$ & $68.61_{[67.73,69.50]}$ & $68.13_{[67.22,69.02]}$ & $68.13_{[67.22,69.02]}$ & $84.46_{[83.75,85.14]}$ & $85.11_{[84.39,85.82]}$ & $85.11_{[84.39,85.82]}$ \\
& passthrough     & $74.55_{[73.84,75.31]}$ & $73.19_{[72.47,73.95]}$ & $73.19_{[72.49,73.92]}$ & $69.94_{[69.06,70.85]}$ & $69.72_{[68.85,70.62]}$ & $69.72_{[68.85,70.62]}$ & $84.81_{[84.11,85.51]}$ & $85.11_{[84.42,85.80]}$ & $85.11_{[84.42,85.80]}$ \\
& task\_arithmetic & $74.59_{[73.86,75.30]}$ & $73.58_{[72.81,74.30]}$ & $73.58_{[72.86,74.30]}$ & $69.81_{[68.93,70.67]}$ & $69.13_{[68.25,70.02]}$ & $69.13_{[68.25,70.02]}$ & $85.07_{[84.38,85.76]}$ & $85.42_{[84.73,86.09]}$ & $85.42_{[84.73,86.09]}$ \\
\addlinespace
\multirow{3}{*}{qwen25}
& linear          & $71.04_{[70.33,71.84]}$ & $70.98_{[70.27,71.75]}$ & $70.98_{[70.22,71.78]}$ & $58.54_{[57.55,59.57]}$ & $58.50_{[57.52,59.53]}$ & $58.50_{[57.52,59.53]}$ & $71.59_{[70.72,72.47]}$ & $71.62_{[70.76,72.50]}$ & $71.62_{[70.76,72.50]}$ \\
& passthrough     & $68.72_{[67.98,69.51]}$ & $68.69_{[67.93,69.48]}$ & $68.69_{[67.98,69.46]}$ & $56.68_{[55.69,57.68]}$ & $56.78_{[55.80,57.78]}$ & $56.78_{[55.80,57.78]}$ & $66.85_{[65.88,67.74]}$ & $66.83_{[65.86,67.72]}$ & $66.83_{[65.86,67.72]}$ \\
& task\_arithmetic & $70.10_{[69.36,70.87]}$ & $70.10_{[69.35,70.85]}$ & $70.10_{[69.39,70.86]}$ & $57.32_{[56.35,58.32]}$ & $57.32_{[56.35,58.32]}$ & $57.32_{[56.35,58.32]}$ & $69.45_{[68.55,70.34]}$ & $69.45_{[68.55,70.34]}$ & $69.45_{[68.55,70.34]}$ \\
\addlinespace
\multirow{3}{*}{smollm2}
& linear          & $49.23_{[48.38,50.00]}$ & $49.18_{[48.33,50.00]}$ & $49.18_{[48.30,49.95]}$ & $52.08_{[51.07,53.10]}$ & $52.05_{[51.05,53.08]}$ & $52.05_{[51.05,53.08]}$ & $67.52_{[66.62,68.46]}$ & $67.43_{[66.52,68.37]}$ & $67.43_{[66.52,68.37]}$ \\
& passthrough     & $49.22_{[48.41,50.04]}$ & $49.22_{[48.41,50.03]}$ & $49.22_{[48.38,50.06]}$ & $52.03_{[51.06,53.02]}$ & $52.03_{[51.06,53.02]}$ & $52.03_{[51.06,53.02]}$ & $66.83_{[65.92,67.75]}$ & $66.83_{[65.92,67.75]}$ & $66.83_{[65.92,67.75]}$ \\
& task\_arithmetic & $50.01_{[49.15,50.79]}$ & $50.01_{[49.15,50.78]}$ & $50.01_{[49.20,50.82]}$ & $52.27_{[51.26,53.28]}$ & $52.27_{[51.26,53.28]}$ & $52.27_{[51.26,53.28]}$ & $67.48_{[66.56,68.41]}$ & $67.48_{[66.56,68.41]}$ & $67.48_{[66.56,68.41]}$ \\
\bottomrule
\end{tabular}
}
\caption{MergeKit suite: full downstream evaluation results (with 95\% CI) on different cases. Arena and in-memory columns are \method{}.}
\label{tab:downstream-mergekit-full}
\end{table*}

\begin{table*}[t]
\centering
\small
\setlength{\tabcolsep}{5pt}
\renewcommand{\arraystretch}{1.15}
\begin{tabular}{l ccc}
\toprule
\textbf{Metric} & \textbf{full\_base} & \textbf{upstream\_rys\_7\_10} & \textbf{brainsurgery\_rys\_7\_10} \\
\midrule
bbh/causal\_judgement [exact\_match]              & $0.1600_{[0.0574,0.2626]}$ & $0.2200_{[0.1040,0.3360]}$ & $0.2200_{[0.1040,0.3360]}$ \\
bbh/date\_understanding [exact\_match]            & $0.6200_{[0.4841,0.7559]}$ & $0.6200_{[0.4841,0.7559]}$ & $0.6200_{[0.4841,0.7559]}$ \\
bbh/logical\_deduction\_five\_objects [exact\_match] & $0.0200_{[0.0000,0.0592]}$ & $0.0000_{[0.0000,0.0000]}$ & $0.0000_{[0.0000,0.0000]}$ \\
bbh/navigate [exact\_match]                       & $0.0400_{[0.0000,0.0949]}$ & $0.0400_{[0.0000,0.0949]}$ & $0.0400_{[0.0000,0.0949]}$ \\
gsm8k\_cot [flexible-extract]                     & $0.7000_{[0.5717,0.8283]}$ & $0.7400_{[0.6172,0.8628]}$ & $0.7400_{[0.6172,0.8628]}$ \\
gsm8k\_cot [strict-match]                         & $0.2800_{[0.1543,0.4057]}$ & $0.3600_{[0.2256,0.4944]}$ & $0.3600_{[0.2256,0.4944]}$ \\
ifeval [inst\_level\_loose\_acc]                   & $0.8553$\textsuperscript{a} & $0.8947$\textsuperscript{a} & $0.8947$\textsuperscript{a} \\
ifeval [inst\_level\_strict\_acc]                  & $0.8158$\textsuperscript{a} & $0.8553$\textsuperscript{a} & $0.8553$\textsuperscript{a} \\
ifeval [prompt\_level\_loose\_acc]                 & $0.7800_{[0.6640,0.8960]}$ & $0.8400_{[0.7374,0.9426]}$ & $0.8400_{[0.7374,0.9426]}$ \\
ifeval [prompt\_level\_strict\_acc]                & $0.7200_{[0.5943,0.8457]}$ & $0.7800_{[0.6640,0.8960]}$ & $0.7800_{[0.6640,0.8960]}$ \\
mbpp [pass\_at\_1]                                 & $0.9000_{[0.8160,0.9840]}$ & $0.8400_{[0.7374,0.9426]}$ & $0.8400_{[0.7374,0.9426]}$ \\
\midrule
\textbf{Average (all metrics)} & \textbf{0.5356} & \textbf{0.5627} & \textbf{0.5627} \\
\bottomrule
\end{tabular}
\caption{LLM Neuroanatomy suite: downstream evaluation results (95\% CI in subscript, computed as value $\pm 1.96\times$ stderr from the \texttt{lm-evaluation-harness} output. \textsuperscript{a}\texttt{ifeval} instruction-level accuracy metrics stderr is not computed by author's evaluation setup.}
\label{tab:downstream-neuro-full}
\end{table*}

\begin{table*}[t]
\centering
\scriptsize
\setlength{\tabcolsep}{5pt}
\renewcommand{\arraystretch}{1.1}
\begin{tabular}{ll rrrrrr}
\toprule
\textbf{Case} & \textbf{System} & \textbf{Total s} & \textbf{GiB/s} & \textbf{Peak PSS/input} & \textbf{Avg PSS/input} & \textbf{Logical I/O} & \textbf{Physical I/O} \\
\midrule
\multirow{3}{*}{block\_l2}
& LLM-Pruner            & 27.758 & 0.452 & 3.346 & 1.803 & 2.805 & 0.807 \\
& \method{} (arena)      & 80.670 & 0.156 & 1.657 & 1.055 & 1.805 & 2.390 \\
& \method{} (in-memory)  & 35.542 & 0.353 & 1.074 & 0.615 & 1.805 & 0.805 \\
\addlinespace
\multirow{3}{*}{block\_taylor}
& LLM-Pruner            & 28.649 & 0.438 & 3.347 & 1.847 & 2.805 & 0.807 \\
& \method{} (arena)      & 77.845 & 0.161 & 1.657 & 1.064 & 1.805 & 2.390 \\
& \method{} (in-memory)  & 34.556 & 0.363 & 1.076 & 0.637 & 1.805 & 0.805 \\
\addlinespace
\multirow{3}{*}{layer\_wise}
& LLM-Pruner            & 19.622 & 0.640 & 2.060 & 0.836 & 2.760 & 0.762 \\
& \method{} (arena)      & 35.598 & 0.353 & 1.157 & 0.763 & 1.760 & 1.760 \\
& \method{} (in-memory)  & 14.272 & 0.879 & 0.819 & 0.285 & 1.760 & 0.760 \\
\bottomrule
\end{tabular}
\caption{LLM-Pruner suite: full run-cost results by case and system. Peak/Avg PSS and Logical/Physical I/O are normalized per input byte; Physical I/O is the combined kernel-reported storage-layer read+write bytes, Logical I/O is the combined tensor-payload read+write bytes.}
\label{tab:runcost-pruner-full}
\end{table*}

\begin{table*}[t]
\centering
\scriptsize
\setlength{\tabcolsep}{4pt}
\renewcommand{\arraystretch}{1.05}
\begin{tabular}{lll rrrrrr}
\toprule
\textbf{Model} & \textbf{Merge type} & \textbf{System} & \textbf{Total s} & \textbf{GiB/s} & \textbf{Peak PSS/input} & \textbf{Avg PSS/input} & \textbf{Logical I/O} & \textbf{Physical I/O} \\
\midrule
\multirow{9}{*}{olmo2}
& \multirow{3}{*}{linear}
  & MergeKit              & 134.284 & 0.894 & 1.007 & 0.523 & 1.500 & 0.500 \\
& & \method{} (arena)      & 247.737 & 0.485 & 1.082 & 0.697 & 1.500 & 2.099 \\
& & \method{} (in-memory)  & 156.743 & 0.766 & 1.006 & 0.504 & 1.500 & 0.500 \\
\cmidrule(l){2-9}
& \multirow{3}{*}{passthrough}
  & MergeKit              & 109.615 & 0.548 & 2.194 & 1.450 & 2.091 & 1.091 \\
& & \method{} (arena)      & 122.963 & 0.488 & 1.157 & 0.826 & 2.091 & 2.182 \\
& & \method{} (in-memory)  & 47.073  & 1.275 & 1.105 & 0.544 & 2.091 & 1.091 \\
\cmidrule(l){2-9}
& \multirow{3}{*}{task\_arithmetic}
  & MergeKit              & 127.473 & 0.942 & 1.006 & 0.537 & 1.500 & 0.500 \\
& & \method{} (arena)      & 227.288 & 0.528 & 1.068 & 0.715 & 1.500 & 2.025 \\
& & \method{} (in-memory)  & 126.653 & 0.948 & 1.007 & 0.578 & 1.500 & 0.500 \\
\midrule
\multirow{9}{*}{qwen25}
& \multirow{3}{*}{linear}
  & MergeKit              & 37.289 & 0.761 & 1.045 & 0.395 & 1.500 & 0.500 \\
& & \method{} (arena)      & 54.248 & 0.523 & 1.156 & 0.602 & 1.500 & 1.506 \\
& & \method{} (in-memory)  & 38.437 & 0.738 & 1.030 & 0.387 & 1.500 & 0.500 \\
\cmidrule(l){2-9}
& \multirow{3}{*}{passthrough}
  & MergeKit              & 18.623 & 0.762 & 2.238 & 0.917 & 2.092 & 1.092 \\
& & \method{} (arena)      & 36.181 & 0.392 & 1.312 & 0.689 & 2.092 & 2.184 \\
& & \method{} (in-memory)  & 16.258 & 0.872 & 1.152 & 0.388 & 2.092 & 1.092 \\
\cmidrule(l){2-9}
& \multirow{3}{*}{task\_arithmetic}
  & MergeKit              & 39.101 & 0.726 & 1.028 & 0.384 & 1.500 & 0.500 \\
& & \method{} (arena)      & 50.429 & 0.563 & 1.151 & 0.587 & 1.500 & 1.503 \\
& & \method{} (in-memory)  & 36.027 & 0.787 & 1.030 & 0.504 & 1.500 & 0.500 \\
\midrule
\multirow{9}{*}{smollm2}
& \multirow{3}{*}{linear}
  & MergeKit              & 11.929 & 0.534 & 2.122 & 0.464 & 1.500 & 0.500 \\
& & \method{} (arena)      & 17.733 & 0.360 & 1.564 & 0.461 & 1.500 & 1.506 \\
& & \method{} (in-memory)  & 19.698 & 0.324 & 1.111 & 0.276 & 1.500 & 0.501 \\
\cmidrule(l){2-9}
& \multirow{3}{*}{passthrough}
  & MergeKit              & 9.512  & 0.335 & 3.397 & 0.593 & 2.078 & 1.078 \\
& & \method{} (arena)      & 13.386 & 0.238 & 2.190 & 0.502 & 2.078 & 2.158 \\
& & \method{} (in-memory)  & 11.091 & 0.287 & 1.344 & 0.295 & 2.078 & 1.079 \\
\cmidrule(l){2-9}
& \multirow{3}{*}{task\_arithmetic}
  & MergeKit              & 10.842 & 0.588 & 2.122 & 0.424 & 1.500 & 0.500 \\
& & \method{} (arena)      & 15.305 & 0.417 & 1.582 & 0.440 & 1.500 & 1.501 \\
& & \method{} (in-memory)  & 15.681 & 0.407 & 1.133 & 0.366 & 1.500 & 0.501 \\
\bottomrule
\end{tabular}
\caption{MergeKit suite: full run-cost results by case and system. Peak/Avg PSS and Logical/Physical I/O are normalized per input byte; Physical I/O is the combined kernel-reported storage-layer read+write bytes, Logical I/O is the combined tensor-payload read+write bytes.}
\label{tab:runcost-mergekit-full}
\end{table*}

\begin{table*}[t]
\centering
\small
\setlength{\tabcolsep}{6pt}
\renewcommand{\arraystretch}{1.1}
\begin{tabular}{l rrrrrr}
\toprule
\textbf{System} & \textbf{Total s} & \textbf{GiB/s} & \textbf{Peak PSS/input} & \textbf{Avg PSS/input} & \textbf{Logical I/O} & \textbf{Physical I/O} \\
\midrule
LLM Neuroanatomy & 44.379  & 1.375 & 1.010 & 0.431 & 2.089 & 1.045 \\
\method{} (arena)         & 110.238 & 0.554 & 1.088 & 0.796 & 2.045 & 2.089 \\
\method{} (in-memory)     & 45.913  & 1.329 & 1.059 & 0.538 & 2.045 & 1.045 \\
\bottomrule
\end{tabular}
\caption{LLM Neuroanatomy suite: full run-cost results by case and system. Peak/Avg PSS and Logical/Physical I/O are normalized per input byte; Physical I/O is the combined kernel-reported storage-layer read+write bytes, Logical I/O is the combined tensor-payload read+write bytes.}
\label{tab:runcost-neuro-full}
\end{table*}

\section{List of Evaluation Benchmarks with references}

The benchmarks used for system comparison (\ref{sec:tool-comparison} are: HellaSwag~\citep{zellers-etal-2019-hellaswag}, BoolQ~\citep{clark2019boolq}, OpenBookQA~\citep{mihaylov2018openbookqa}, ARC-E/C~\citep{clark2018arc}, WinoGrande~\citep{sakaguchi2021winogrande}, PIQA~\citep{bisk2020piqa}, MMLU~\citep{hendrycks2021mmlu}, BBH~\citep{suzgun2022bbh}, GSM8K~\citep{cobbe2021gsm8k}, IFEval~\citep{zhou2023ifeval}, MBPP~\citep{austin2021mbpp}.

\end{document}